%% file: main.tex
\documentclass{article} 
\usepackage{iclr2023_conference,times}
\input{math_commands}

\usepackage{url}
\usepackage{import}

\usepackage{graphicx,epstopdf}
\graphicspath{{imgs/}}
\usepackage{caption}
\usepackage{subcaption}
\usepackage{diagbox}

 \usepackage[T1]{fontenc}    

\usepackage{amsfonts}       
\usepackage{amsthm}
\usepackage{algorithm}
\usepackage{algpseudocode}
\usepackage{cleveref}
\newtheorem{theorem}{Theorem}[section]
\newtheorem{lemma}{Lemma}[section]
\newtheorem{assumption}{Assumption}[section]
\newtheorem{remark}{Remark}[section]
\newtheorem{definition}{Definition}[section]
\newtheorem{example}{Example}[section]

\usepackage{tikz}
\usetikzlibrary{shapes, shadows, arrows}
\usetikzlibrary{positioning}
\usepackage{verbatim}
\tikzstyle{block} = [draw, rectangle, 
    minimum height=3em, minimum width=6em]
\tikzstyle{sum} = [draw, circle, node distance=1cm]
\tikzstyle{input} = [coordinate]
\tikzstyle{output} = [coordinate]
\tikzstyle{pinstyle} = [pin edge={to-,thin,black}]

\title{Backstepping Temporal Difference Learning}

\iclrfinalcopy

\author{Han-Dong Lim  \\
Department of Electrical Engineering \\
KAIST, Daejeon, 34141, South Korea \\
\texttt{limaries30@kaist.ac.kr} \\
\And
Donghwan Lee  \\
Department of Electrical Engineering \\
KAIST, Daejeon, 34141, South Korea \\
\texttt{donghwan@kaist.ac.kr} \\
}

\begin{document}

\maketitle

\begin{abstract}
        \import{introduction}{abstract.tex}
\end{abstract}
 \section{Introduction}
     \import{introduction}{intro.tex}
\section{Preliminaries}
    \import{preliminaries}{main}
\section{Designing TD-learning through backstepping}\label{sec:btd}
    \import{algorithms}{main}


\section{Experiments}
    \import{./}{experiments.tex}
\section{Conclusion}
    \import{./}{conclusion.tex}
\section{Acknowledgements}
    \import{./}{acknowledgements.tex}
\bibliography{biblio}
\bibliographystyle{iclr2023_conference}

\clearpage
\section{Appendix}
\import{appendix}{main}

\newpage

\end{document}

%% file: math_commands.tex
\usepackage{amsmath,amsfonts,bm}

\def\eqref#1{equation~\ref{#1}}

\def\1{\bm{1}}

\DeclareMathAlphabet{\mathsfit}{\encodingdefault}{\sfdefault}{m}{sl}
\SetMathAlphabet{\mathsfit}{bold}{\encodingdefault}{\sfdefault}{bx}{n}



%% file: introduction/abstract.tex
 Off-policy learning ability is an important feature of reinforcement learning (RL) for practical applications. However, even one of the most elementary RL algorithms, temporal-difference (TD) learning, is known to suffer form divergence issue when the off-policy scheme is used together with linear function approximation. To overcome the divergent behavior, several off-policy TD-learning algorithms, including gradient-TD learning (GTD), and TD-learning with correction (TDC), have been developed until now. In this work, we provide a unified view of such algorithms from a purely control-theoretic perspective, and propose a new convergent algorithm. Our method relies on the backstepping technique, which is widely used in nonlinear control theory.
 Finally, convergence of the proposed algorithm is experimentally verified in environments where the standard TD-learning is known to be unstable.

%% file: introduction/intro.tex
Since~\cite{mnih2015human}, which has demonstrated that deep reinforcement learning (RL) outperforms human in several video games (Atari 2600 games), significant advances has been made in RL theory and algorithms. For instance,~\cite{van2016deep,lan2020maxmin,chen2021randomized} proposed some variants of the so-called deep Q-network~\citep{mnih2015human} that achieves higher scores in Atari games than the original deep Q-network. An improved deep RL was developed in~\cite{badia2020agent57} that performs better than average human scores across 57 Atari games. Not only performing well in video games, but~\cite{schrittwieser2020mastering} also have shown that an RL agent can self-learn chess, Go, and Shogi. Furthermore, RL has shown great success in real world applications, e.g., robotics~\citep{kober2013reinforcement}, healthcare~\citep{gottesman2019guidelines}, and recommendation systems~\citep{chen2019generative}.
\par
 Despite the practical success of deep RL, there is still a gap between theory and practice. One of the notorious phenomena is the deadly triad~\citep{sutton2018reinforcement}, the diverging issue of the algorithm when function approximation, off-policy learning, and bootstrapping are used together. One of the most fundamental algorithms, the so-called temporal-difference (TD) learning~\citep{sutton1988learning}, is known to diverge under the deadly triad, and several works have tried to fix this issue for decades. In particular, the seminar works~\cite{sutton2008convergent,sutton2009fast} introduced the so-called GTD, gradient-TD2 (GTD2), and TDC, which are off-policy, and have been proved to be convergent with linear function approximation. More recently,~\cite{ghiassian2020gradient} suggested regularized version of TDC called TD learning with regularized correction (TDRC), and showed its favorable features under off-policy settings. Moreover,~\cite{lee2021versions} developed several variants of GTD based on primal dual formulation.
\par
 On the other hand, backstepping control~\citep{khalil2015nonlinear} is a popular method in designing stable controllers for nonlinear systems with special structures. The design technique offers a wide range of stable controllers, and is proved to be robust under various settings. It has been used in various fields including quadrotor helicopters~\citep{madani2006backstepping}, mobile robots~\citep{fierro1997control}, and ship control~\citep{fossen1999tutorial}. Using backstepping control technique, in this paper, we develop a new convergent off-policy TD-learning which is a single time-scale algorithm. 
\par
In particular, the goal of this paper is to introduce a new unifying framework to design off-policy TD-learning algorithms under linear function approximation. The main contributions are summarized as follows:
\begin{itemize}
     \item We propose a systemic way to generate off-policy TD-learning algorithms including GTD2 and TDC from control theoretic perspective. 
     \item Using our framework, we derive a new TD-learning algorithm, which we call backstepping TD (BTD).
     \item We experimentally verify its convergence and performance under various settings including where off-policy TD has known to be unstable.
 \end{itemize}
 
 In particular, most of the previous works on off-policy TD-learning algorithms (e.g., GTD2 and TDC) are derived based on optimization perspectives starting with an objective function. Then, the convergence is proved by proving stability of the corresponding O.D.E. models. In this paper, we follow reversed steps, and reveal that an off-policy TD-learning algorithm (called backstepping TD) can be derived based on control theoretic motivations. In particular, we develop stable O.D.E. models first using the backstepping technique, and then recover back the corresponding off-policy TD-learning algorithms. The new analysis reveals connections between off-policy TD-learning and notions in control theory, and provides additional insights on off-policy TD-learning with simple concepts in control theory. This sound theoretical foundation established in this paper can potentially motivate further analysis and developments of new algorithms.

\par
Finally, we briefly summarize TD learning algorithms that guarantee convergence under linear function approximation. GTD~\citep{sutton2008convergent}, GTD2 and TDC~\citep{sutton2009fast} have been developed to approximate gradient on mean squared projected Bellman error. Later, GTD and GTD2 has been discovered to solve minimax optimization problem~\citep{macua2014distributed,liu2020finite}. Such sadde-point view point of GTD has led to many interesting results including~\cite{du2017stochastic,dai2018sbeed,lee2021versions}. TDRC~\citep{ghiassian2020gradient} adds an additional term similar to regularization term to one-side of parameter update, and tries to balance between the performance of TD and stability of TDC. TDC++~\citep{ghiassian2020gradient} also adds an additional regularization term on both sides of the parameter update. Even though TDRC shows good performance, it uses additional parameter condition to ensure convergence, whereas TDC++ does not.

%% file: preliminaries/main.tex
\subsection{Nonlinear system theory}
\import{./preliminaries}{nonlinear.tex}

\subsection{Stochastic approximation and O.D.E. approach}

\import{./preliminaries}{borkar.tex}
\subsection{Backstepping control}\label{sec:backstepping}
\import{./preliminaries}{back.tex}
\subsection{Markov Decision Process}
\import{./preliminaries}{mdp.tex}
\subsection{Temporal difference learning}
 \import{./preliminaries}{td.tex}

\subsection{Gradient temporal difference learning}
 \import{./preliminaries}{gtd.tex}

%% file: preliminaries/nonlinear.tex
Nonlinear system theory will play an important role throughout this paper.
Here, we briefly review basics of nonlinear systems. Let us consider the continuous-time nonlinear system
\begin{align}
    \dot{x}_t =f(x_t,u_t) ,\quad x_0 \in\mathbb{R}^n,\label{eq:nonlinear-system}
\end{align}
where \(x_0\in\mathbb{R}^n\) is the initial state, \(t\in\mathbb{R},t\geq 0\) is the time, \(x_t\in\mathbb{R}^n\) is the state, \(u_t\in\mathbb{R}^n\) is the control input, and \(f:\mathbb{R}^n\times \mathbb{R}^n \to \mathbb{R}^n \) is a nonlinear mapping. 
An important concept in dealing with nonlinear systems is the equilibrium point. Considering the state-feedback law $u_t = \mu(x_t)$, the system can be written as $\dot{x}_t =f(x_t,u_t)=f(x_t,\mu(x_t))=:f(x_t)$, and a point $x=x^e$ in the state-space is said to be an equilibrium point of~(\ref{eq:nonlinear-system}) if it has the property that whenever the state of the system starts at $x^e$, it will remain at $x^e$~\citep{khalil2015nonlinear}. For~$\dot{x}_t =f(x_t)$, the equilibrium points are the real roots of the equation $f(x)=0$. The equilibrium point $x^e$ is said to be globally asymptotically stable if for any initial state $x_0 \in {\mathbb R}^n$, $x_t \to x^e$ as $t \to \infty$. 

An important control design problem is to construct a state-feedback law $u_t = \mu(x_t)$ such that the origin becomes the globally asymptotically stable equilibrium point of~(\ref{eq:nonlinear-system}).
To design a state-feedback law to meet such a goal, control Lyapunov function plays a central role, which is defined in the following definition.
\begin{definition}[Control Lyapunov function~\citep{sontag2013mathematical}]
A positive definite function \(V:\mathbb{R}^n\to\mathbb{R}\) is called a control Lyapunov function (CLF) if for all \(x \neq 0\), there exists a corresponding control input \(u \in {\mathbb R}^m\) that satisfies the inequality, \( \nabla_x V(x)^{\top} f(x,u)<0\) for all $x \neq 0$.
\end{definition}
Once such a CLF is found, then it guarantees that there exists the control law that stabilizes the system.
Moreover, the corresponding state-feedback control law can be extracted from the CLF, e.g., $\mu (x) = \arg \min _u \nabla_x V(x)^{\top} f(x,u)$ provided that the minimum exists and unique. The concept of control Lyapunov function will be used in the derivations of our main results. For the autonomous system, $\dot{x}_t =f(x_t)$, and Lypaunov function $V: {\mathbb R}^n \to {\mathbb R}$, Lie derivative is defined as ${\cal L}_f V(x): =  \nabla_x V(x)^{\top} f(x)$ so that $\dot V(x_t ) = {\cal L}_f V(x_t )$ along the solution.

%% file: preliminaries/borkar.tex
Including Q-learning~\citep{watkins1992q} and TD-learning~\citep{sutton1988learning}, reinforcement learning algorithms can be considered as stochastic approximation~\citep{robbins1951stochastic} described by
\begin{align}
    x_{k+1} = x_k + \alpha_k ( f(x_k) + \epsilon_k), \label{eq:sa}
\end{align}
where \( f: \mathbb{R}^n \to \mathbb{R}^n \) is a nonlinear mapping, and \( \epsilon_k\) is an i.i.d. noise. Borkar and Meyn theorem~\citep{borkar2000ode} is a well-known method to bridge the asymptotic convergence of stochastic approximation and the stability of its corresponding O.D.E. model, which can be expressed as
\begin{align}
    \dot{x}_t = f(x_t),\quad x_0 \in\mathbb{R}^n, \label{eq:nonlinear_ode}
\end{align}
where \(x_0 \in \mathbb{R}^n\) is initial state, and $t \in \mathbb{R}$, $t \geq 0$ is the time.

Borkar and Meyn theorem~\citep{borkar2000ode} states that under the conditions in~\Cref{assm:borkar_meyn} in the Appendix, global asymptotic stability of the O.D.E.~(\ref{eq:nonlinear_ode}) leads to asymptotic convergence of the stochastic approximation update~(\ref{eq:sa}), which is formally stated in the following lemma.
\begin{lemma}[Borkar and Meyn theorem~\citep{borkar2000ode}]\label{borkar_meyn_lemma}
    Suppose that~\Cref{assm:borkar_meyn} in the Appendix holds, and consider the stochastic approximation in~(\ref{eq:sa}). Then, for any initial \(x_0 \in \mathbb{R}^n \), \(\sup_{k\geq 0}||x_k|| < \infty \) with probability one. In addition , 
    \( x_k \rightarrow x^e \) as \( k \rightarrow \infty \) with probability one, where \(x^e\) is the unique equilibrium point of the O.D.E. in~(\ref{eq:nonlinear_ode}).
\end{lemma}

The main idea of Borkar and Meyn theorem is as follows: iterations of a stochastic recursive algorithm follow the solution of its corresponding O.D.E. in the limit when the step-size satisfies the so-called Robbins-Monro condition~\citep{robbins1951stochastic} in~(\ref{cond:robbins_monro_step_size}) in the Appendix. Therefore, by proving asymptotic stability of the O.D.E., we can induce convergence of the original algorithm. In this paper, we will use an O.D.E. model of TD-learning, which is expressed as a linear time-invariant system. 

%% file: preliminaries/back.tex
This section provides the concept of the backstepping control~\citep{kokotovic1992joy,khalil2015nonlinear}, which will be the main tool in this paper to derive TD-learning algorithms. The backstepping technique is a popular tool for generating a CLF (control Lyapunov function) for nonlinear systems with specific structures. In particular, let us start with the following general nonlinear system:
\begin{align}
    \dot{y}_t &= f(y_t) + g(y_t) x_t \label{eq:ode:prelim}\\
    \dot{x}_t &= u_t \nonumber,
\end{align}
where \( y_t \in\mathbb{R}^m, x_t \in\mathbb{R}^m\) are the states, \(u_t \in\mathbb{R}^m \) is the input, and \( f:\mathbb{R}^m \to \mathbb{R}^m\) and \(g:\mathbb{R}^m \to\mathbb{R}\) are continuous functions. The first system is a nonlinear system with a particular affine structure, and the second system is simply an integrator. It can be seen as a cascade interconnection of two systems, where the second system's state is injected to the input of the first system. The backstepping control technique gives us a systematic way to generate a CLF for such particular nonlinear systems provided that the first system admits a CLF independently. To this end, we suppose that the first system admits a CLF. Through the backstepping approach, designing a stable control law for the above system can be summarized in the following steps:
\begin{itemize}
\item[Step 1.] Consider \(x_t\) in~(\ref{eq:ode:prelim}) as virtual input \(\tilde{x}(y_t)\) (state-feedback controller), and consider the following system:
\(
    \dot{\lambda}_t = f(y_t) + g(y_t) \tilde{x}(y_t) \nonumber
\).
Design \(\tilde{x}(y_t)\) such that the above system admits a CLF $V$, i.e., it admits a positive definite and radially unbounded function \(V\) such that its time derivative is negative definite, i.e.,\( \dot{V}(y_t) < 0, \forall y_t \neq 0 \).
\item[Step 2.] Denote the error between the virtual state-feedback controller \(\tilde{x}(y_t)\) and state variable \(x_t\) as \(z_t:=  x_t- \tilde{x}(y_t) \nonumber\). Now, rewrite the original O.D.E. in~(\ref{eq:ode:prelim}) with the new variable \((y_t,z_t)\):
\(\frac{d}{dt}\begin{bmatrix}y_t\\z_t\end{bmatrix} = \begin{bmatrix}
f(y_t) +g(y_t)\tilde{x}(y_t) + g(y_t) z_t\\
 u_t -\dot{\tilde{x}}(y_t)
\end{bmatrix}\)
\item[Step 3.] Design the control input \(u_t\) such that the above system is stable. One popular choice is to consider the CLF \(
   V_c(y_t,z_t) := V(y_t) +||z_t||^2/2 
\), where \(V(y_t)\) is defined in Step 1. Then choose \(u_t\) such that the time derivative of \(V_c(y_t,z_t) \) to be negative definite. 
\end{itemize}



A simple example of designing stabilizing control law by backstepping technique is given in Appendix~\Cref{app:ex_back}.

%% file: preliminaries/mdp.tex
In this paper, we consider a Markov decision process (MDP) characterized by the tuple \( (\mathcal{S},\mathcal{A},\mathcal{P},\gamma,r)\), where \( \mathcal{S}:=\{1,2,\dots,|\mathcal{S}|\}\) stands for the set of finite state space, $|\mathcal{S}|$ denotes the size of $\mathcal{S}$, \(\mathcal{A}:=\{1,2,\dots,|\mathcal{A}|\} \) denotes the set of finite action space, $|\mathcal{A}|$ is the size of $\mathcal{A}$, \(\gamma\in(0,1)\) is the discount factor, \(\mathcal{P}:\mathcal{S}\times\mathcal{A}\times\mathcal{S}\to[0,1]\) denotes the Markov transition kernel, and \(r:\mathcal{S}\times\mathcal{A}\times\mathcal{S}\to\mathbb{R}\) means the reward function. In particular, if an agent at state \(s\in\mathcal{S}\), takes action \(a\in\mathcal{A}\), then the current state transits to the next state \(s^{\prime}\in\mathcal{S}\) with probability \(\mathcal{P}(s,a,s^{\prime}) \), and the agent receives reward \( r(s,a,s^{\prime}) \). Each element of the state to state transition matrix under policy \(\pi\), denoted by \(P^{\pi}\in\mathbb{R}^{|\mathcal{S}|\times |\mathcal{S}|}\) is \(
        [P^{\pi}]_{ij}:=
   \sum\limits_{a\in\mathcal{A}} \pi(a|i)\mathcal{P}(i,a,j),\quad 1\leq i,j\leq |\mathcal{S}| \nonumber,
\)
where \([P^{\pi}]_{ij}\) corresponds to \(i\)-th row and \(j\)-th column element of matrix \(P^{\pi}\). Moreover, the stationary state distribution induced by policy \(\mu\), is denoted as \( d^{\mu}:\mathcal{S}\to[0,1]\), i.e., \(d^{\mu\top}P^{\mu}=d^{\mu\top} \). With the above setup, we define the following matrix notations: 
\begin{align}
    D^{\mu} &:=\begin{bmatrix}
    d^{\mu}(1)&&\\
    & \ddots &\\
    && d^{\mu}(|\mathcal{S}|)
    \end{bmatrix} 
    \in\mathbb{R}^{|\mathcal{S}|\times |\mathcal{S}|} ,\quad
        R^{\pi} = \begin{bmatrix}
    \mathbb{E}_{a\sim \pi}[r(s,a,s^{\prime})|s=1]\\
    \mathbb{E}_{a\sim \pi}[r(s,a,s^{\prime})|s=2]\\
    \vdots\\
    \mathbb{E}_{a\sim \pi}[r(s,a,s^{\prime})|s=|\mathcal{S}|]\\
    \end{bmatrix} \in\mathbb{R}^{|\mathcal{S}|} \nonumber,
\end{align}
where \(D^{\mu}\) is a diagonal matrix of the state distribution induced by behavior policy \(\mu\), each element of \(R^{\pi}\) is the expected reward under policy \(\pi\) at the corresponding state. The policy evaluation problem aims to approximate the value function at state \(s\in\mathcal{S}\), $v^\pi  (s): = {\mathbb E}\left[ \left. {\sum_{k = 0}^\infty  {\gamma ^k } r(S_k ,A_k ,S_{k + 1} )} \right|S_0  = s,\pi \right]$, where the trajectory is generated under policy \(\pi:\mathcal{S}\times\mathcal{A}\to [0,1]\). In this paper, we consider the linear function approximation to approximate the value function \(v^{\pi}(s)\). In particular, we parameterize the value function \(v^{\pi}(s)\) with $\phi^{\top}(s)\xi$, where \( \phi:\mathcal{S}\to \mathbb{R}^{n}\) is a pre-selected feature vector with $\phi(s):=\begin{bmatrix} \phi_1(s) &\cdots & \phi_n(s) \end{bmatrix}$, $\phi_1,\ldots,\phi_n:{\mathcal S}\to {\mathbb R}$ are feature functions, and \(\xi\in\mathbb{R}^n\) is the learning parameter. The goal of the policy evaluation problem is then to approximate the value function $v^{\pi}(s)$ using this linear parameterization, i.e., $\phi^{\top}(s)\xi\approx v^{\pi}(s)$. Moreover, using the matrix notation \(
    \Phi := 
    \begin{bmatrix}
    \phi(1),
    \phi(2),
    \cdots,
    \phi(|\mathcal{S}|)
    \end{bmatrix}^{\top} \in\mathbb{R}^{|\mathcal{S}|\times n},
\) called the feature matrix, the linear parameterization can be written in the vector form $\Phi \xi$. 
We also assume that \(\Phi\) is full column rank matrix throughout the paper, which is a standard assumption~\citep{sutton2008convergent,sutton2009fast,ghiassian2020gradient,lee2021versions}.

%% file: preliminaries/td.tex
This section provides a brief background on TD-learning~\citep{sutton1988learning}. Suppose that we have access to stochastic samples of state \(s_k\) from the state stationary distribution induced by the behavior policy \(\mu\), i.e., \( s_k \sim d^{\mu}(\cdot) \), and action is chosen under behavior policy \(\mu\), i.e., \(a_k \sim \mu(\cdot|s_k)\). Then, we observe the next state \(s^{\prime}_k\) following \( s^{\prime}_k \sim \mathcal{P}(\cdot,a_k,s_k) \), and receive the reward \(r_k := r(s_k,a_k,s^{\prime}_k) \). Using the simplified notations for the feature vectors \(
    \phi_k := \phi(s_k), \quad \phi^{\prime}_k = \phi(s^{\prime}_k) \nonumber.
\)
the TD-learning update at time step $k$ with linear function approximation can be expressed as \(
 \xi_{k+1} = \xi_k + \alpha_k \rho_k \delta_k(\xi_k) \phi_k,
\) where $\alpha_k >0$ is the step-size, $\delta_k(\xi_k) := r_k + \gamma \phi^{\prime\top}_k \xi_k - \phi^{\top}_k\xi_k$ is called the temporal difference or temporal difference error (TD-error), and $\rho_k := \rho(s_k,a_k)=\frac{\pi(a_k|s_k)}{\mu(a_k|s_k)}$ is called the importance sampling ratio~\citep{precup2001off}. The importance sampling ratio re-weights the TD-error to handle the mismatch between the behavior policy \(\mu\) and target policy \(\pi\). It is known that TD-learning with linear function approximation and off-policy learning scheme does not guarantee convergence in general. The above stochastic approximation aims to find fixed point of the following projected Bellman equation, which is, after some manipulations, expressed as:
\begin{align}
    \Phi^{\top}D^{\mu}\Phi \xi^* - \gamma \Phi^{\top}D^{\mu}P^{\pi}\Phi \xi^* = \Phi^{\top}D^{\mu}R^{\pi}.\label{eq:pbe}
\end{align}

To simplify the expressions, let use introduce one more piece of notations:
\begin{align}
    A &:= \mathbb{E}_{s\sim d^{\mu}(s),s^{\prime}\sim P^{\pi}(s^{\prime}|s)}[\phi(s)(\phi(s)-\gamma \phi(s^{\prime}))^{\top}] =\Phi^{\top}D^{\mu}\Phi-\gamma \Phi^{\top}D^{\mu}P^{\pi}\Phi \in \mathbb{R}^{n\times n},\nonumber\\
    b &:= \mathbb{E}_{s\sim d^{\mu}(s),a\sim\pi(a|s),s^{\prime}
    \sim P(s^{\prime}|s,a) }[r(s,a,s^{\prime})\phi(s)] = \Phi^{\top}D^{\mu} R^{\pi} \in\mathbb{R}^{n \times 1}.\nonumber
\end{align}

Even though we can use arbitrary distribution, for simplicity we assume stationary distribution of \(\mu\). 
Now, we can rewrite~(\ref{eq:pbe}) compactly as
\begin{align}
    A \xi^* = b  \label{eq:td_fixed_point} .
\end{align}

The corresponding O.D.E. for TD-learning can be written as $\dot{\xi}_t = A\xi_t -b, \xi_0 \in {\mathbb R}^n$. Using the coordinate transform \(x_k := \xi_k -\xi^*\), we get the O.D.E. $\dot{x}_t = Ax_t, x_0 \in {\mathbb R}^n$, whose origin is globally asymptotically stable equilibrium point if $\rho(s,a)=\frac{\pi(a|s)}{\mu(a|s)} = 1$ for all $(s,a) \in {\cal S} \times {\cal A}$.
Throughout the paper we will use the vector \( x_k := \xi_k -\xi^*\) to represent the coordinate transform of \(\xi_k\) to the origin, and will use \( \xi_t \) and \(x_t\) to denote the corresponding continuous-time counterparts of \( \xi_k \) and \(x_k\), respectively.

%% file: preliminaries/gtd.tex
To fix the instability issue of off-policy TD-learning under linear function approximation,~\cite{sutton2008convergent} and~\cite{sutton2009fast} introduced various stable off-policy TD-learning algorithms, called GTD (gradient TD-learning), GTD2, and TDC (temporal difference correction). The idea behind these algorithms is to minimize the mean-square error of  projected Bellman equation (MSPBE) $\min_{\xi\in\mathbb{R}^n} \frac{1}{2} ||\Phi^{\top}D^{\mu}(R^{\pi}+\gamma P^{\pi}\Phi\xi -\Phi\xi )||^2_{(\Phi^{\top}D^{\mu}\Phi)^{-1}}$, where \( ||x||_{D}:= \sqrt{x^{\top}Dx}\), and the global minimizer of MSPBE corresponds to the solution of~(\ref{eq:td_fixed_point}). The core idea of the algorithms is to introduce an additional variable \(\lambda_k \in \mathbb{R}^n\) to approximate the stochastic gradient descent method for MSPBE as an objective function. In particular, GTD2 update can be written as
\begin{align}
    \lambda_{k+1} = \lambda_k + \alpha_k ( -\phi^{\top}_k \lambda_k +\rho_k \delta_k(\xi_k))\phi_k ,\quad 
    \xi_{k+1} = \xi_k + \alpha_k ( \phi_k^{\top}\lambda_k \phi_k -\rho_k \gamma \phi^{\top}_k\lambda_k\phi_k^{\prime}) \nonumber.
\end{align}
We denote \( \lambda_t\) to denote continuous time part of \(\lambda_k \). Since the fixed point for \(\lambda_k \) is zero, it doesn't require coordinate transformation. It is a single time-scale algorithm because it uses a single step-size $\alpha_k$. The corresponding O.D.E. is expressed as $\dot{\lambda}_t = - C\lambda_t - Ax_t ,\dot{x}_t = A^{\top}\lambda_t$, where \(
        C := \mathbb{E}_{s\sim d^{\mu}(s)}[\phi(s)\phi^{\top}(s)]=\Phi^{\top}D^{\mu}\Phi \in \mathbb{R}^{n\times n}.
\)
Similarly, TDC update can be written as
\begin{align}
        \lambda_{k+1} &= \lambda_k + \alpha_k ( -\phi^{\top}_k \lambda_k +\rho_k \delta_k(\xi_k))\phi_k \label{eq:tdc_faster}\\
    \xi_{k+1} &= \xi_k + \beta_k ( -\rho_k \gamma \phi^{\top}_k\lambda_k\phi_k^{\prime} + \rho_k \delta_k(\xi_k)\phi_k) \label{eq:tdc_slower},
\end{align}
where the step-sizes, \(\alpha_k\) and \(\beta_k\), satisfy \( \alpha_k /\beta_k \to 0 \) as \(k\to\infty\) and the Robbins and Monro step-size condition~\citep{robbins1951stochastic} in~(\ref{cond:robbins_monro_step_size}) in Appendix. It is a two time-scale algorithm because it uses two time-steps, $\alpha_k$ and $\beta_k$.



%% file: algorithms/main.tex
\import{algorithms}{motivation}

\subsection{Backstepping TD}\label{subsec:btd}
\import{algorithms}{btd}

\subsection{Recovering single time-scale TDC}\label{sec:ss_tdc}
    \import{./}{single_time_tdc.tex}
\subsection{Generalizing TDC++}

\import{./algorithms}{tdc_plus.tex}

%% file: algorithms/motivation.tex
We briefly explain the motivation for our algorithmic development. Borkar and Meyn theorem~\citep{borkar2000ode} in~\Cref{borkar_meyn_lemma} is a typical tool to prove convergence of Q-learning~\citep{borkar2000ode,lee2019unified} and TD-learning~\citep{sutton2009fast,lee2021versions}.
Most of the previous works on off-policy TD-learning algorithms (e.g., GTD2 and TDC) first start with an objective function, and then derive GTD algorithms based on optimization perspectives. Then, the convergence is proved using the corresponding O.D.E. models and stability theory of linear time-invariant systems. A natural question arises is, can we derive off-policy TD-learning algorithms following a reversed step? In other words, can we develop a stable O.D.E. model first using tools in control theory, and then recover back the corresponding off-policy TD-learning algorithms? In this paper, we reveal that a class of off-policy TD-learning algorithms can be derived based on purely control theoretic motivations following such a reversed process. By doing so, this work provides additional insights on off-policy TD-learning algorithms and gives a sound theoretical foundation on off-policy TD-learning algorithms for further developments of new algorithms.

Designing stabilizing control laws for continuous-time nonlinear system has been successful over the past decades~\citep{khalil2015nonlinear}. One such technique, so called backstepping, is a popular controller design method in non-linear control literature~\citep{khalil2015nonlinear}. With the help of the backstepping method~\citep{khalil2015nonlinear}, we design stabilizing control laws for continuous-time systems, and then the corresponding off-policy TD-learning algorithms are derived, and are shown to be convergent via Borkar and Meyn theorem~\citep{borkar2000ode} in~\Cref{borkar_meyn_lemma}. The brief procedure is explained in the following steps: Step 1) Choose an appropriate continuous-time dynamic model such that
    (a) we can recover the TD-fixed point \(\xi^*\) in~(\ref{eq:td_fixed_point}) via its equilibrium point;
    (b) the corresponding stochastic approximation algorithm can be implementable only through transitions of MDP and accessible data.;
    Step 2) Using the backstepping method, design a control input to stabilize the dynamic model chosen in Step 1).

%% file: algorithms/btd.tex
Now, we introduce a new off-policy TD-learning algorithm, which we call Backstepping TD (BTD). Firstly, we will develop a stabilizing control law for the following the continuous-time system:
\begin{align}
    \dot{\lambda}_t &= (-C + \eta A ) \lambda_t - Ax_t\label{sys:lambda} \\
    \dot{x}_t &= u_t  \label{sys:x}
\end{align}
The idea stems from finding a control system for which we can easily apply the backstepping techinque. In details, the backstepping techinqiue can be applied to the two interconnected systems where one subsystem, namely~(\ref{eq:ode:prelim}), can be stabilized with \(x_t \) in~(\ref{eq:ode:prelim}) as a control input. 
Therefore, our first aim is to find such a system. 
To this end, we can try a natural choice of O.D.E. to solve the TD problem, i.e., $\dot \lambda_t = A \lambda_t$, which is however unstable in the off-policy case. 
Therefore, we can develop a modified O.D.E. $\dot \lambda_t = (- C +  \eta A) \lambda_t - A x_t$, where $x_t$ is the control input, the negative definite matrix $-C$ is introduced to stabilize the system, and $\eta >0$ is introduced to provide additional degrees of freedom in design. Now, the constructed system can be stabilized through the state-feedback controller $x_t = \eta \lambda_t$ and admits the simple control Lypaunov function $V(\lambda) = ||\lambda ||^2$. Moreover, $A$ should be included in the right-hand side in order to implement the corresponding algorithm without knowing the solution because $x _k = \xi _k  - \xi ^*$ and $\xi ^*$ should be removed using $A\xi ^*  = b$ in the final step. Simply setting $x_t = \eta \lambda_t$ may cancel out $A$ in the right-hand side, the O.D.E. becomes $\dot{\lambda}_t = - C\lambda_t$, Therefore, as mentioned before, we can apply the backstepping technique by adding an additional dynamic controller. As the next step, the backstepping technique is applied, and one needs to observe what would be the final form of the control system. In summary, if we consist \(f(\lambda_t)\) with the combination of \(A \) and \( -C\)  (not necessarily \(-C\), it may be \(-I\)) , it can be a reasonable candidate to apply the backstepping technique. Cancelling \( A\) with virtual input only leaves \( -C\), which guarantees stability from its negative definiteness. Therefore,~(\ref{sys:lambda}) and~(\ref{sys:x}) is a reasonable candidate for the dynamics where we can apply the backstepping technique. 
 In particular, our aim is to design an appropriate control input \(u_t\) for the above system such that the origin is the unique asymptotically stable equilibrium point, i.e., \( (\lambda_t,x_t) \to 0 \) as $t \to \infty$ for any $(\lambda_0,x_0) \in {\mathbb R}^n \times {\mathbb R}^n$. The overall procedure is depicted in~\Cref{fig:btd} in the Appendix, and we show how to choose the control input \(u_t\) in the following lemma.
\begin{lemma}\label{lem:btd_ode}
Consider the O.D.E. in~(\ref{sys:lambda}) and~(\ref{sys:x}). If we choose the control input \( u_t :=  (A^{\top} + \eta^2A -\eta  C )\lambda_t- \eta Ax_t\), then the above O.D.E. has globally asymptotically stable origin, i.e.,
\( (\lambda_t,x_t) \to (0,0)\) as \(t\to \infty\) for any $(\lambda_0,x_0) \in {\mathbb R}^n \times {\mathbb R}^n$.
\end{lemma}
\begin{proof}[Proof sketch]

The proof follows the steps given in the backstepping scheme in~\Cref{sec:btd}.
First, substituting \(x_t\) in~(\ref{sys:lambda}) with a virtual controller \(\tilde{x}(\lambda_t)\), we will design a control law \( \tilde{x} (\lambda_t ) \) that stabilizes the following new virtual system:
\begin{align}
\dot{\lambda}_t = (-C+\eta A) \lambda_t - A\tilde{x}(\lambda_t). \label{eq:virtual}
\end{align}
One natural choice of the virtual controller is \(\tilde{x}(\lambda_t) = \eta \lambda_t\). Plugging it into~(\ref{eq:virtual}) leads to \(
    \dot{\lambda}_t = - C \lambda_t
\), and we can verify the global asymptotic stability of the above system with the following Lyapunov function:
\begin{align}
    V (\lambda_t) := \frac{||\lambda_t||_2^2}{2} . \label{eq:V_1}
\end{align}
We now consider the original O.D.E. in~(\ref{sys:lambda}) and~(\ref{sys:x}). Applying simple algebraic manipulations yield \( 
   \dot{\lambda}_t = - C \lambda_t - A (x_t-\eta \lambda_t ) ,
  \quad
    \dot{x}_t = u_t 
\).
The error between \(x_t\) and the virtual controller \(\tilde{x}(\lambda_t)\) can be expressed as new variable \(z_t\), which is \(
   z_t:= x_t-\tilde{x}(\lambda_t) = x_t- \eta \lambda_t 
\). Rewriting the O.D.E. in~(\ref{sys:lambda}) and~(\ref{sys:x}) with \((\lambda_t, z_t)\) coordinates, we have

\begin{align}
    \dot{\lambda}_t &= -C \lambda_t - Az_t \label{eq:sys:lambda:btd} \\
    \dot{z}_t &= u_t + \eta  C \lambda_t +  \eta Az_t. \nonumber
\end{align}
To prove the global asymptotic stability of the above system, consider the function \(V_c(\lambda_t, z_t): = V(\lambda_t) + \frac{1}{2}||z_t||^2_2\) where \( V(\lambda_t)\) is defined in~(\ref{eq:V_1}). By taking \(u_t\) as $u_t = A^{\top}\lambda_t -\eta C\lambda_t  - \eta  Az_t $, we can apply LaSall'es invariance principle in~\Cref{lem:lasalle}. The full proof is in Appendix~\Cref{app:lem:btd_ode}.
\end{proof}

Using the relation $z_t:= x_t- \eta \lambda_t$, the control input in the original coordinate $(\lambda_t ,x_t)$ can be written as
\(
    u_t := A^{\top}\lambda_t -\eta C\lambda_t  - \eta  Az_t =  (A^{\top} + \eta^2A- \eta C)\lambda_t - \eta Ax_t.
\)
Plugging this input into the original open-loop system in~(\ref{sys:lambda}) and~(\ref{sys:x}), the closed-loop system in the original coordinate $(\lambda_t ,x_t)$ can written as
\begin{align}
   \dot{\lambda}_t &= ( -C  +\eta A ) \lambda_t - Ax_t \label{btd:ode_lambda}\\
    \dot{x}_t &= (A^{\top} + \eta^2A -\eta  C )\lambda_t- \eta Ax_t, \label{btd:ode_x}
\end{align}
whose origin is also globally asymptotically stable according to~\Cref{lem:btd_ode}. Recovering back from \(x_t\) to \(\xi_t\), we have
\( \frac{d}{dt}\begin{bmatrix}
\lambda_t\\
\xi_t
\end{bmatrix} = 
\begin{bmatrix}
-C+\eta A & -A\\
A^{\top}+\eta^2 A -\eta C & -\eta A
\end{bmatrix}
\begin{bmatrix}
\lambda_t\\
\xi_t
\end{bmatrix} + \begin{bmatrix}
b\\
\eta b
\end{bmatrix} 
\).
The corresponding stochastic approximation of the O.D.E. in~\Cref{thm:btd} becomes
\begin{align}
    \lambda_{k+1} &= \lambda_k +\alpha_k  (( (-1+\eta)\phi^{\top}_k - \eta \rho_k \gamma \phi^{\prime\top}_k )\lambda_k  +\rho_k\delta_k(\xi_k) )\phi_k  \label{btd:lambda_update}\\
    \xi_{k+1} &= \xi_k +\alpha_k(((-\eta+\eta^2)\phi^{\top}_k - \eta^2  \rho_k \gamma \phi^{\prime\top}_k )\lambda_k \phi_k + \eta \rho_k \delta_k(\xi_k)\phi_k  + ( \phi^{\top}_k\lambda_k \phi_k - \rho_k \gamma  \phi^{\top}_k \lambda_k \phi^{\prime}_k )   ) \label{btd:xi_update}.
\end{align}

The equilibrium point of the above O.D.E. is \((0,\xi^*)\). Hence, we only need to transform the coordinate of \(\xi_t\) to \(x_t = \xi_t - \xi^* \), which results to the O.D.E. in~(\ref{btd:ode_lambda}) and~(\ref{btd:ode_x}). With the above result, we are now ready to prove convergence of~\Cref{algo:btd}. The proof simply follows from Borkar and Meyn theorem in~\Cref{borkar_meyn_lemma}, of which the details can be found in~\cite{sutton2009fast}.
\begin{theorem}\label{thm:btd}
Under the step size condition~(\ref{cond:robbins_monro_step_size}) , with~\Cref{algo:btd} in Appendix, \(\xi_k \to \xi^*\) as \( k \to \infty \) with probability one, where \(\xi^*\) is the fixed point of~(\ref{eq:td_fixed_point}).
\end{theorem}
\begin{proof}
The proof is done by checking~\Cref{assm:borkar_meyn} in Appendix.
\end{proof}

\begin{remark}
~\Cref{thm:btd} doesn't require any condition on \(\eta\). Therefore, we can set \(\eta =0 \), which results to GTD2 developed in~\cite{sutton2009fast}.
\end{remark}


%% file: single_time_tdc.tex
In this section, we derive a single-time scale version of TDC~\citep{sutton2009fast} through the backstepping design in the previous section. TDC~\citep{sutton2009fast} was originally developed as a two-time scale algorithm in~\cite{sutton2009fast}. Even though the two time-scale method provides theoretical guarantee for a larger class of algorithms, the single time-scale scheme provides more simplicity in practice, and shows faster convergence empirically. Subsequently, \cite{maei2011gradient} provided a single-time scale version of TDC by multiplying a large enough constant $\eta >0$ to the faster time scale part~(\ref{eq:tdc_faster}), which leads to
\begin{align}
  \lambda_{k+1} &= \lambda_k + \beta_k \eta (- \phi^{\top}_k\lambda_k +   \rho_k\delta_k(\xi_k))\phi_k  \label{tdc:lambda_update}\\
    \xi_{k+1} &= \xi_k+ \beta_k  (-\rho_k \gamma \phi^{\top}_k\lambda_k \phi_k^{\prime} + \rho_k \delta_k(\xi_k)\phi_k
    )  \label{tdc:xi_update},
\end{align}
where
\begin{align}
\eta>\max\left\{0,-\lambda_{\min}\left(C^{-1}(A+A^{\top})/2\right)\right\}\label{ineq:meai_eta}.
\end{align}Here, we derive another version of single-time TDC by multiplying a constant to the slower time-scale part in~(\ref{eq:tdc_slower}), which results in
\begin{align}
  \lambda_{k+1} &= \lambda_k + \alpha_k (- \phi^{\top}_k\lambda_k +  \rho_k \delta_k(\xi_k))\phi_k \label{eq:new_tdc_faster}  \\
    \xi_{k+1} &= \xi_k+ \alpha_k \beta  (\phi^{\top}_k \lambda_k \phi_k-  \rho_k \gamma \phi^{\top}_k \lambda_k \phi^{\prime}_k +\rho_k\delta_k(\xi_k)\phi_k
    ) , \label{eq:new_tdc_slower}
\end{align}
where \(\beta\) satisfies
\begin{equation}\label{cond:beta}
        0<   \beta < -\frac{\lambda_{\min}(C)}{\lambda_{\min}(A)} \quad \text{if} \quad \lambda_{\min} (A) < 0 ,\; \text{else}\quad \beta >0.
\end{equation}
We can derive the above algorithm following similar steps as in~\Cref{subsec:btd}. Let us first consider the following dynamic model:
\begin{align}
    \dot{\lambda}_t &= -C \lambda_t - Ax_t \label{eq:tdc_ode}\\
    \dot{x}_t &= u_t   \label{eq:tdc_ode_2}
\end{align}
Using the backstepping technique, we can prove that the above system admits the origin as a global asymptotically stable equilibrium point with the control input \( u_t:=\beta \left((A^{\top}-C) \lambda_t - A \xi_t \right) \), which is shown in the following lemma:
\begin{lemma}\label{lem:st_tdc}
Consider the O.D.E. in~(\ref{eq:tdc_ode}) and~(\ref{eq:tdc_ode_2}).
Suppose that we choose the control input \(   u_t:=\beta \left((A^{\top}-C) \lambda_t - A \xi_t \right)) \), and \(\beta\) satisfies condition~(\ref{cond:beta}). Then, the above O.D.E. has globally asymptotically stable origin, i.e., \( (\lambda_t,x_t) \to (0,0)\) as \(t\to \infty\).
\end{lemma}
The proof of~\Cref{lem:st_tdc} is given in Appendix~\Cref{app:lem:st_tdc}. By Borkar and Meyn theorem in~\Cref{borkar_meyn_lemma}, we can readily prove the convergence of~\Cref{algo:tdc} in Appendix, which uses stochastic recursive update~(\ref{eq:new_tdc_faster}) and~(\ref{eq:new_tdc_slower}).

\begin{theorem}\label{thm:tdc_slow}
Consider~\Cref{algo:tdc} in Appendix. Under the step size condition~(\ref{cond:robbins_monro_step_size}), and if \(\beta\) satisfies~(\ref{cond:beta}), \(\xi_k \to \xi^*\) as \( k \to \infty \) with probability one, where \(\xi^*\) is the fixed point of~(\ref{eq:td_fixed_point}).
\end{theorem}

We will call the~\Cref{algo:tdc2} as TDC-slow, and single-time version of TDC suggested by~\cite{maei2011gradient} as TDC-fast. Other than the multiplication of a constant reflecting two-time scale property, we can make TDC into a single-time algorithm, which we call a single time-scale TDC2, while the original version in~\cite{maei2011gradient} will be called the single time-scale TDC. The derivation is given in Appendix~\Cref{app:tdc2}. The performance of such versions of TDC are evaluated in Appendix~\Cref{app:tdc_exp}. Even though not one of the algorithms outperforms each other, TDC-slow and TDC2 shows better performance in general. 

%% file: algorithms/tdc_plus.tex
This section provides versions of TDC++~\citep{ghiassian2020gradient}, which is variant of TDC. With an additional regularization term \(\xi_k\) on both updates of TDC in~(\ref{eq:tdc_faster}) and~(\ref{eq:tdc_slower}), the update is written as follows:
\begin{align}
  \lambda_{k+1} &= \lambda_k + \alpha_k  \eta(- \phi^{\top}_k\lambda_k + \rho_k \delta_k(\xi_k))\phi_k- \beta \lambda_k )  \label{tdc++:lambda_update}\\
    \xi_{k+1} &= \xi_k+ \alpha_k  (- \rho_k \gamma \phi^{\top}_k \lambda_k \phi^{\prime}_k-\beta \lambda_k + \rho_k \delta_k(\xi_k)\phi_k
    )  \label{tdc++:xi_update},
\end{align}
where \( \eta >0 \) satisfies~(\ref{ineq:meai_eta}) and \(\beta >0\) is a new parameter. Note that TDC++ can be simply viewed as variant of TDC by adding the term $\beta \lambda_k$ in the update, which can be seen as a regularization term. Therefore, letting $\beta = 0$ yields the original TDC. In this paper, we prove that our controller design leads to the following update:
\begin{align}
  \lambda_{k+1} &= \lambda_k + \alpha_k  \eta(- \phi^{\top}_k\lambda_k +   \rho_k \delta_k(\xi_k))\phi_k- \beta \lambda_k )  \label{eq:tdc++2:lambda_update}\\
    \xi_{k+1} &= \xi_k+ \alpha_k  (-\rho_k \gamma \phi^{\top}_k\lambda_k \phi^{\prime}_k+(1-\kappa \eta)\phi_k^{\top}\lambda_k\phi_k - \kappa \beta \eta \lambda_k + \rho_k \kappa \eta\delta_k(\xi_k)\phi_k
    )  \label{eq:tdc++2:xi_update},
\end{align}
where \(\kappa \) and \(\beta\) are new parameters and when \(\kappa = 1/\eta \) it becomes TDC++.
The difference with the original TDC++ can be seen in their corresponding O.D.E. forms.
The corresponding O.D.E. for~(\ref{tdc++:lambda_update}) and~(\ref{tdc++:xi_update}) (original TDC++) can be expressed as:
\(\frac{d}{dt}
\begin{bmatrix}
    \lambda_t\\
    x_t
\end{bmatrix}
=\begin{bmatrix}
    -\eta (C+\beta I) & -\eta A\\
    A^{\top}-C -\beta I & -A
\end{bmatrix}
\begin{bmatrix}
    \lambda_t\\
    x_t
\end{bmatrix}
\). Meanwhile, the O.D.E. corresponding to~(\ref{eq:tdc++2:lambda_update}) and~(\ref{eq:tdc++2:xi_update}) (new TDC++) becomes
\(
\frac{d}{dt}\begin{bmatrix}
    \lambda_t\\
    x_t
\end{bmatrix}=
\begin{bmatrix}
      -\eta ( C + \beta I) & - \eta A \\
      A^{\top}- \kappa \eta(C + \beta I) & -\kappa \eta A
\end{bmatrix}\begin{bmatrix}
      \lambda_t\\
      x_t
\end{bmatrix}
\). We experiment under different of \( \kappa \) and \(\eta\) to examine the behavior of new TDC++. The result shows that in general, smaller \( \kappa  \) leads to better performance. The results are given in Appendix~\Cref{app:exp_tdc++}.

\begin{lemma}\label{lem:tdc++ode}
Consider the following O.D.E.:
\begin{align}
    \dot{\lambda}_t &= -\eta (C+ \beta I) \lambda_t -\eta Ax_t \label{eq:tdc+_ode}\\
    \dot{x}_t &= u_t  .\label{eq:tdc+u}
\end{align}
Suppose that we choose the control input \(   u_t:=  (A^{\top}- \kappa \eta (C+\beta I)) \lambda_t - \kappa \eta A x_t  \). Assume \(\eta>0\) and \(\beta\) and \(\kappa\) satisfies the following condition:
\(
    \beta + \kappa \lambda_{\min} (A)  > \lambda_{\min} (C) \label{ineq:beta_tdc++}.
\)
Then, the above O.D.E. has globally asymptotically stable origin, i.e.,
\( (\lambda_t,x_t) \to (0,0)\) as \(t\to \infty\).
\end{lemma}
The proof is given in Appendix~\Cref{app:lem:tdc++ode}. With~\Cref{borkar_meyn_lemma}, we can prove the convergence of stochastic update with~(\ref{eq:tdc++2:lambda_update}) and~(\ref{eq:tdc++2:xi_update}) whose pseudo code is given in~\Cref{algo:tdc++2} in Appendix.
\begin{theorem}\label{thm:tdc++2}
Consider~\Cref{algo:tdc++2} in Appendix. Under the step-size condition~(\ref{cond:robbins_monro_step_size}) and if \(\eta\) satisfies~(\ref{ineq:meai_eta}), then \(\xi_k \to \xi^*\) as \( k \to \infty \) with probability one, where \(\xi^*\) is the TD fixed point in~(\ref{eq:td_fixed_point}).
\end{theorem}

\begin{remark}
We can replace the regularization term with nonlinear terms satisfying certain conditions. The details are given in Appendix~\Cref{app:tdc++nonlinear}.
\end{remark}

%% file: experiments.tex
We verify the performance and convergence of the proposed BTD under standard benchmarks to evaluate off-policy TD-learning algorithms, including Baird environment~\citep{baird1995residual}, RandomWalk~\citep{sutton2009fast} with different features, and Boyan chain~\citep{boyan2002technical}. The details about the environments are given in Appendix~\Cref{app:exp_env}. From the experiments, we see how BTD behaves under different coefficients \( \eta \in \{-0.5,-0.25,0,0.25,0.5\} \). We measure the Root Mean-Squared Projected Bellman Error (RMSPBE) as the performance metric, and every results are averaged over 100 runs. From~\Cref{table:btd}, the result with \(\eta = 0.5\) shows the best performance except at Baird, where \(\eta=0\), corresponding to GTD2 performs best. There exist two aspects on the role of \( \eta \). First of all, it can be thought of as a parameter that can mitigate the effect of instability coming from matrix \( A \) in~(\ref{sys:lambda}). For example, a smaller $\eta$ can stabilize the system. However, as a trade off, if \( \eta \) is too small, then the update rate might be too small as well. As a result, the overall convergence can be slower. Furthermore, \( \eta \) also controls the effect of \( -C \) in~(\ref{eq:sys:lambda:btd}) in the BTD update rules, where $-C$ corresponds to \((-\eta+\eta^2)\phi_k^{\top}\lambda_k\phi_k\) in~(\ref{btd:xi_update}). Note that the role of \( \eta \) in the final BTD update rule in~(\ref{btd:xi_update}) shows different perspectives compared to that in~(\ref{sys:lambda}). In particular, \( \eta = 1/2 \) maximizes the effect of \( -C \) in~(\ref{btd:xi_update}). From~\Cref{table:btd}, it leads to reasonably good performances in most domains.
Another natural choice is to multiply \(\eta \) to \(-C \) instead of \( A\). However, in such cases, we need to introduce another constrain \( \eta>0 \), whereas in the current BTD, convergence is guaranteed for all \(\eta \in \mathbb{R} \). Finally, we note that simply multiplying \( - C\) by a large positive constant does not lead to good results in general. This is because in this case, it may increase variance, and destabilize the algorithm. Overall results are given in Appendix~\Cref{app:exp_btd}.

\begin{table}[H]
\centering
\caption{Backstepping TD, step-size = $0.01$}\label{table:btd}
\begin{tabular}{|c|c|c|c|c|c|}\hline
\backslashbox{Env}{$\eta$}
&\makebox[1em]{-0.5}&\makebox[1em]{-0.25}&\makebox[1em]{0}
&\makebox[1em]{0.25}&\makebox[1em]{0.5}\\\hline\hline
Boyan &$1.51 \pm 0.66$ &$1.481 \pm 0.656$ &$1.452 \pm 0.647$ &$1.428 \pm 0.64$ &${\color{blue}1.408 \pm 0.635}$\\\hline
Dependent &$0.11 \pm 0.19$ &$0.097 \pm 0.163$ &$0.086 \pm 0.142$ &$0.079 \pm 0.128$ &${\color{blue}0.076 \pm 0.122}$\\\hline
Inverted &$0.21 \pm 0.25$ &$0.173 \pm 0.218$ &$0.151 \pm 0.193$ &$0.139 \pm 0.177$ &${\color{blue}0.136 \pm 0.172}$\\\hline
Tabular &$0.17 \pm 0.28$ &$0.147 \pm 0.238$ &$0.133 \pm 0.208$ &$0.124 \pm 0.191$ &${\color{blue}0.122 \pm 0.188}$\\\hline
Baird &$0.1 \pm 0.64$ &$0.09 \pm 0.629$ &${\color{blue}0.085 \pm 0.625}$ &$0.087 \pm 0.628$ &$0.092 \pm 0.637$\\\hline
\end{tabular}
\hfill
\end{table}

%% file: conclusion.tex
In this work, we have proposed a new framework to design off-policy TD-learning algorithms from control-theoretic view. Future research directions would be extending the framework to non-linear function approximation setting.


%% file: acknowledgements.tex
This work was supported by the National Research Foundation under Grant  NRF-2021R1F1A1061613, Institute of Information communications Technology Planning Evaluation (IITP) grant funded by the Korea government (MSIT)(No.2022-0-00469), and the BK21 FOUR from the Ministry of Education (Republic of Korea). (Corresponding author: Donghwan Lee.)

%% file: appendix/main.tex
\subsection{Techinical details}
\import{appendix}{control}
\subsection{Omiitted pseudo codes and diagrmas}
\import{appendix}{algorithm}
\import{appendix}{btd_diagram}
\subsection{Example of backstepping}\label{app:ex_back}
\import{appendix}{example_backstepping}

\subsection{Omitted Proofs}
\import{appendix}{proof}
\subsection{Derivation of single time-scale TDC2}\label{app:tdc2}
 \import{algorithms}{tdc2.tex}

\subsection{TDC++ with nonlinear terms}\label{app:tdc++nonlinear}
\import{algorithms}{nonlinear.tex}

\import{appendix/experiments}{main}

\import{appendix}{experiments}
\subsection{Background on GTD2 and TDC}
\import{./appendix}{background_additional}

%% file: appendix/control.tex
We elaborate the conditions for the Borkar and Meyn Theorem~\citep{borkar2000ode}. Consider the stochastic approximation in~(\ref{eq:sa}).
\begin{assumption}\label{assm:borkar_meyn}
1. The mapping \(f: \mathbb{R}^n \rightarrow \mathbb{R}^n\) is globally Lipschitz continuous, and there exists a function \(f_{\infty} : \mathbb{R}^n \rightarrow \mathbb{R}^n\) such that
    \begin{equation}
        \lim_{c\rightarrow\infty} \frac{f(cx)}{c} = f_{\infty}(x) , \quad \forall{x} \in \mathbb{R}^n.    
    \end{equation}
2. The origin in \(\mathbb{R}^n\) is an asymptotically stable equilibrium for the O.D.E.: \(\dot{x}_t = f_{\infty}(x_t)\).
\newline
\newline
3. There exists a unique globally asymptotically stable equilibrium \(x^e\in\mathbb{R}^n\) for the ODE \( \dot{x}_t=f(x_t)\) , i.e., \(x_t \rightarrow x^e\) as \( t\rightarrow\infty\).
\newline
\newline
4. The sequence \(\{ m_k , k\geq 1  \}  \) where \( \mathcal{G}_k \) is sigma-algebra generated by \(\{(x_i,m_i,i\geq k ) \}\), is a Martingale difference sequence. In addition , there exists a constant \( 
C_0 < \infty \) such that for any initial \(  x_0 \in \mathbb{R}^n \) , we have \(\mathbb{E}[|| m_{k+1} ||^2 | \mathcal{G}_k ] \leq C_0 (1+|| x_k ||^2), \forall{k}\geq 0  \).
\newline
\newline
5. The step-sizes satisfies the Robbins-Monro condition~\citep{robbins1951stochastic} :
\begin{align}
    \sum\limits_{k=0}^{\infty } \alpha_k = \infty, \quad   \sum\limits_{k=0}^{\infty } \alpha_k^2 < \infty. \label{cond:robbins_monro_step_size}
\end{align}
\end{assumption}

Furthermore, we introduce an important tool to prove stability of O.D.E..
\begin{lemma}[LaSall'es Invariance Principle~\citep{khalil2015nonlinear}]\label{lem:lasalle}
Let the origin be an equilibrium point for~(\ref{eq:nonlinear_ode}). Let \( V: \mathbb{R}^n \to \mathbb{R}\) be a continuously differentiable and positive definite function satisfying the below conditions:
\begin{enumerate}
    \item \(V(x)\) is radially unbounded function , i.e., $||x||\to \infty$ implies $V(x) \to \infty$, 
    \item Consider the Lie derivative ${\cal L}_f V(x): =  \nabla_x V(x)^{\top} f(x)$ so that $\dot V (x_t ) = {\cal L}_f V(x_t )$ along the solution, and it is negative semi-definite, i.e., \({\cal L}_f V(x) \leq 0 \) for all \( x \in \mathbb{R}^n \).
\end{enumerate}
Let \( S := \{x\in\mathbb{R}^n \mid {\cal L}_f V(x)=0 \} \), and suppose that no solution can stay identically in \( S\) other than trivial solution $x \equiv 0$, where we say that a solution stays identically in $S$ if $x(t) \in S, \forall t \geq 0$ . Then, the origin is globally asymptotically stable.
\end{lemma}

\begin{definition}[Invariant set~\citep{khalil2015nonlinear}]
A set M is an invariant set with respect to \( \dot{x} = f(x)\) if \( x_0 \in M \rightarrow x_t \in M \) for all \( t \geq 0 \).
\end{definition}

%% file: appendix/algorithm.tex
\begin{algorithm}[H]
\caption{Backstepping TD}
  \begin{algorithmic}[1]
    \State Initialize $\xi_0,\lambda_0 \in {\mathbb R}^{n}$.
    \State Set the step-size $(\alpha _k )_{k = 0}^\infty$, and the behavior policy $\mu$.
    \For{iteration $k=0,1,\ldots$}
        \State Sample $s_k\sim d^{\mu}$ and $a_k \sim \mu$
        \State Sample $s_k'\sim P(s_k,a_k,\cdot)$ and $r_{k+1}= r(s_k,a_k,s_k')$
        \State Update \(\lambda_k\) and \(\xi_k\) using~(\ref{btd:lambda_update}) and~(\ref{btd:xi_update}) respectively    
        \EndFor
  \end{algorithmic}\label{algo:btd}
\end{algorithm}

\begin{algorithm}[H]
\caption{TDC-slow}
  \begin{algorithmic}[1]
    \State Initialize $\xi_0,\lambda_0 \in {\mathbb R}^{n}$.
    \State Set the step-size $(\alpha _k )_{k = 0}^\infty$, and the behavior policy $\mu$.
    \For{iteration $k=0,1,\ldots$}
        \State Sample $s_k\sim d^{\mu}$ and $a_k \sim \mu$
        \State Sample $s_k'\sim P(s_k,a_k,\cdot)$ and $r_{k+1}= r(s_k,a_k,s_k')$
        \State Update \(\lambda_k\) and \(\xi_k\) using~(\ref{tdc:lambda_update}) and~(\ref{tdc:xi_update}) respectively  \EndFor
  \end{algorithmic}\label{algo:tdc}
\end{algorithm}

\begin{algorithm}[H]
\caption{TDC++2}
  \begin{algorithmic}[1]
    \State Initialize $\xi_0,\lambda_0 \in {\mathbb R}^{n}$.
    \State Set the step-size $(\alpha _k )_{k = 0}^\infty$, and the behavior policy $\mu$.
    \For{iteration $k=0,1,\ldots$}
        \State Sample $s_k\sim d^{\mu`}$ and $a_k \sim \mu$
        \State Sample $s_k'\sim P(s_k,a_k,\cdot)$ and $r_{k+1}= r(s_k,a_k,s_k')$
        \State Update \(\lambda_k\) and \(\xi_k\) using~(\ref{eq:tdc++2:lambda_update}) and~(\ref{eq:tdc++2:xi_update}) respectively  \EndFor
  \end{algorithmic}\label{algo:tdc++2}
\end{algorithm}

%% file: appendix/btd_diagram.tex
\begin{center}
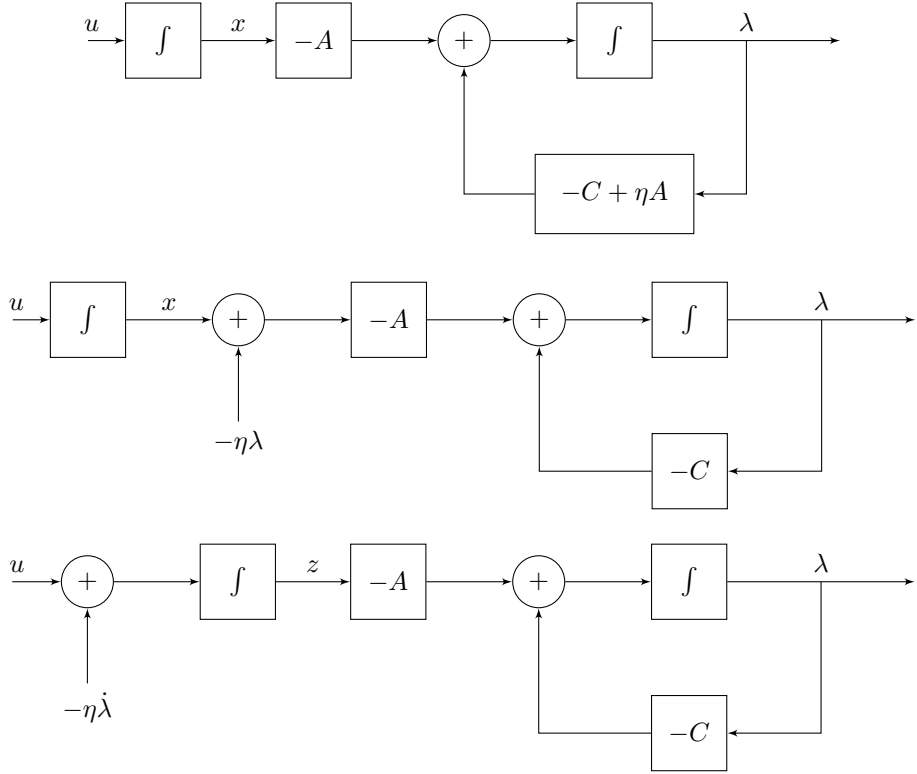
\begin{figure}[H]
  \begin{subfigure}[b]{\textwidth}
  \begin{center}
\begin{tikzpicture}[auto, node distance=3cm,>=latex']
    \node [input, name=input,node distance=1cm] {};
    \node [block, right of=input,minimum height = 1cm,minimum width = 1cm,node distance=1cm] (integral) {$\int$};
    \node [block, right of=integral,minimum height = 1cm,minimum width = 1cm,node distance=2cm] (-A) {$-A$};
    \node [sum, right of=-A,node distance=2cm] (dual_sum){$+$};
    \node [block, right of=dual_sum,minimum height = 1cm,minimum width = 1cm,node distance=2cm] (dual_int) {$\int$};
    \node [output,right of = dual_int] (dual_output){};
    \node [block, below= 1cm of dual_int] (-C+A) {$-C+\eta A$};
    
    \draw [->] (input) -- node[pos=0.1] {$u$}(integral);
    \draw [->] (integral) -- node {$x$} (-A);
    \draw [->] (-A) --  (dual_sum);
    \draw [->] (dual_sum) -- (dual_int);
    \draw [->] (dual_int.east) -- node[pos=0.5,name=output_lambda]{$\lambda$} (dual_output);
    \draw [->] (output_lambda) |-  (-C+A);
    \draw [->] (-C+A) -| (dual_sum);
\end{tikzpicture}
\end{center}
\end{subfigure}
 \par\bigskip 
\begin{center}
  \begin{subfigure}[b]{\textwidth}
  \begin{center}
    \begin{tikzpicture}[auto, node distance=3cm,>=latex']
    \node [input, name=input,node distance=1cm] {};
    \node [block, right of=input,minimum height = 1cm,minimum width = 1cm,node distance=1cm] (integral) {$\int$};
    \node [sum, right of=integral,node distance=2cm] (dual_back) {$+$};
    \node[below= 1cm of dual_back] (dual_input) {$-\eta\lambda$};
    \node [block, right of=dual_back,minimum height = 1cm,minimum width = 1cm,node distance=2cm] (-A) {$-A$};
    \node[sum,right of=-A,node distance=2cm](back_output){$+$};
    \node [block, right of=back_output,minimum height = 1cm,minimum width = 1cm,node distance=2cm] (dual_int) {$\int$};
    \node [output,right of = dual_int] (dual_output){};
    \node [block, below= 1cm of dual_int,minimum height = 1cm,minimum width = 1cm] (-C+A) {$-C$};
    
    \draw [->] (input) -- node[pos=0.1] {$u$}(integral);
    \draw [->] (integral) -- node {$x$} (dual_back);
    \draw [->] (dual_input) -- (dual_back);
    \draw [->] (dual_back) -- (-A);
        \draw [->] (-A) -- (back_output);
    \draw [->] (back_output) -- (dual_int);
    \draw [->] (dual_int.east) -- node[pos=0.5,name=output_lambda]{$\lambda$} (dual_output);
    \draw [->] (output_lambda) |-  (-C+A);
    \draw [->] (-C+A) -| (back_output);
\end{tikzpicture}
  \end{center}
\end{subfigure}%
\par\bigskip 
  \begin{subfigure}[b]{\textwidth}
  \begin{center}
\begin{tikzpicture}[auto, node distance=3cm,>=latex']
    \node [input, name=input,node distance=1cm] {};
    \node [sum, right of=input,node distance=1cm] (dual_back) {$+$};
    \node [block, right of=dual_back,minimum height = 1cm,minimum width = 1cm,node distance=2cm] (integral) {$\int$};
    \node[below= 1cm of dual_back] (dual_input) {$-\eta\dot{\lambda}$};
    \node [block, right of=integral,minimum height = 1cm,minimum width = 1cm,node distance=2cm] (-A) {$-A$};
    \node[sum,right of=-A,node distance=2cm](back_output){$+$};
    \node [block, right of=back_output,minimum height = 1cm,minimum width = 1cm,node distance=2cm] (dual_int) {$\int$};
    \node [output,right of = dual_int] (dual_output){};
    \node [block, below= 1cm of dual_int,minimum height = 1cm,minimum width = 1cm] (-C+A) {$-C$};
    
    \draw [->] (input) -- node[pos=0.1] {$u$}(dual_back);
    \draw [->] (dual_back) -- (integral);
    \draw [->] (integral) -- node {$z$} (-A);
    \draw [->] (dual_input) -- (dual_back);
    \draw [->] (-A) -- (back_output);
    \draw [->] (back_output) -- (dual_int);
    \draw [->] (dual_int.east) -- node[pos=0.5,name=output_lambda]{$\lambda$} (dual_output);
    \draw [->] (output_lambda) |-  (-C+A);
    \draw [->] (-C+A) -| (back_output);
\end{tikzpicture}
\end{center}
\end{subfigure}
\end{center}
    \hfill
    \caption{Backstepping diagram}\label{fig:btd}
\end{figure}
\end{center}

%% file: appendix/example_backstepping.tex
Here, we provide a simple example to design control law using backstepping control.
\begin{example}
Consider the following two-dimensional system:
\begin{align}
    \dot{\lambda}_t & = -\lambda_t^3 - \lambda_t + x_t \label{ex:back_lambda_org}\\
    \dot{x}_t &= u_t \nonumber,
\end{align}
where \( \lambda_t,x_t \in\mathbb{R}\) are the states, and \(u_t\in\mathbb{R}\) is control input.
First, considering \(x_t\) in~(\ref{ex:back_lambda_org}) as virtual input \(x_s(\lambda_t)\), it is easy to check that \(x_s(\lambda_t) = \lambda_t \) satisfies the condition in Step 1 in~\Cref{sec:backstepping}. Substituting \(x_t\) in~(\ref{ex:back_lambda_org})  with \(x_s(\lambda_t) \), we have
\begin{align}
    \dot{\lambda}_t &= -\lambda_t^3 \nonumber.
\end{align}
The globally asymptotically stability of the above system cab be established with the following Lyapunov function:
\begin{align}
    V(\lambda_t) = \frac{\lambda_t^2}{2} \nonumber.
\end{align}
 Let \(z_t:=x_t-x_s(\lambda_t) \). Expressing the O.D.E. in~(\ref{ex:back_lambda_org}) with \( (\lambda_t,z_t)\), we have
\begin{align}
    \dot{\lambda}_t & = -\lambda_t^3  + z_t  \nonumber\\
    \dot{x}_t &= u_t +\lambda_t^3 - z_t  \nonumber.
\end{align}

Suppose we choose a candidate Lyapunov function:
\begin{align}
    V_c(\lambda_t,z_t) = V(\lambda_t) + \frac{z_t^2}{2} \nonumber
\end{align}
The time derivative of \(    V_c(\lambda_t,z_t) \) becomes
\begin{align}
    \dot{V}_c(\lambda_t,z_t) = -\lambda_t^4 +\lambda_tz_t + z_t(u_t+\lambda_t^3 - z_t)  \nonumber.
\end{align}
To make the time derivative negative definite, we can design the control law as:
\begin{align}
    u_t &:= - \lambda_t - \lambda_t^3 \nonumber,
\end{align}
which leads to the following inequality:
\begin{align*}
    \dot{V}_c(\lambda_t,z_t) \leq -\lambda^4_t - z_t^2 
\end{align*}
Now, we can conclude that the origin of the system becomes globally asymptotically stable.
\end{example}

%% file: appendix/proof.tex
\subsubsection{Proof of~\Cref{lem:btd_ode}}\label{app:lem:btd_ode}
\begin{proof}

In this proof, we follow the steps given in the backstepping scheme in~\Cref{sec:btd}.
First, substituting \(x_t\) in~(\ref{sys:lambda}) with a virtual controller \(\tilde{x}(\lambda_t)\), we will design a control law \( \tilde{x} (\lambda_t ) \) that stabilizes the following new virtual system:

\begin{align}
\dot{\lambda}_t = (-C+\eta A) \lambda_t - A\tilde{x}(\lambda_t). \label{eq:app:virtual}
\end{align}
Even though matrix \(C\) is positive definite, due to matrix \(A\), the system may be unstable. One natural choice of the virtual controller is \(\tilde{x}(\lambda_t) = \eta \lambda_t\). Plugging into~(\ref{eq:app:virtual}) leads to \(
    \dot{\lambda}_t = - C\lambda_t
\). The system now has a globally asymptotically stable origin due to the positive definiteness of matrix \(C\). It is straightforward to verify the global asymptotic stability of the above system with the following Lyapunov function:
\begin{align}
    V (\lambda_t) := \frac{||\lambda_t||_2^2}{2} . \label{eq:app:V_1}
\end{align}
With this result in mind, we now consider the original O.D.E. in~(\ref{sys:lambda}) and~(\ref{sys:x}). Applying simple algebraic manipulations yield \( 
   \dot{\lambda}_t = - C \lambda_t - A (x_t-\eta \lambda_t ) ,
  \quad
    \dot{x}_t = u_t 
\).
The error between \(x_t\) and the virtual controller \(\tilde{x}(\lambda_t)\) can be expressed as new variable \(z_t\), which is \(
   z_t:= x_t-\tilde{x}(\lambda_t) = x_t- \eta \lambda_t 
\). Rewriting the O.D.E. in~(\ref{sys:lambda}) and~(\ref{sys:x}) with \((\lambda_t, z_t)\) coordinates, we have
\begin{align}
    \dot{\lambda}_t &= -C \lambda_t - Az_t \label{eq:app:sys:lambda_coordinate_transform}\\
    \dot{z}_t &= u_t + \eta  C \lambda_t +  \eta Az_t. \nonumber
\end{align}

To prove the global asymptotic stability of the above system, consider the function \(V_c(\lambda_t, z_t): = V(\lambda_t) + \frac{1}{2}||z_t||^2_2\) where \( V(\lambda_t)\) is defined in~(\ref{eq:app:V_1}) . The time derivative of the Lyapunov function along the system's solution becomes
\begin{align*}
    \dot{V}_c (\lambda_t, z_t)  &= \lambda^{\top}_t( -C \lambda_t - Az_t) + z^{\top}_t(u_t + \eta C\lambda_t +  \eta Az_t ) \\
             &= -||\lambda_t||^2_C + z^{\top}_t (- A^{\top}\lambda_t +u_t + \eta C\lambda_t+\eta Az_t) .
\end{align*}
By taking \(u_t\) as $u_t = A^{\top}\lambda_t -\eta C\lambda_t  - \eta  Az_t $, the corresponding closed-loop system is $\frac{d}{{dt}}\left[ {\begin{array}{*{20}c}
   {\lambda _t }  \\
   {z_t }  \\
\end{array}} \right] = \left[ {\begin{array}{*{20}c}
   { - C} & { - A}  \\
   {A^{\top} } & 0  \\
\end{array}} \right]\left[ {\begin{array}{*{20}c}
   {\lambda _t }  \\
   {z_t }  \\
\end{array}} \right] = :f(\lambda _t ,z_t )$, and we have \(\dot{V}_c=-||\lambda_t||^2_C \leq 0 \) for all $(\lambda _t ,z_t ) \ne (0,0)$. Since the inequality is not strict, Lyapunov theory cannot be directly applied. Therefore, we will use LaSall'es invariance principle in~\Cref{lem:lasalle} in Appendix. Define the Lie derivative ${\cal L}_f V(\lambda ,z): =  - \left\| \lambda  \right\|_C^2$ so that $\dot V_c (\lambda_t ,z_t ) = {\cal L}_f V(\lambda _t ,z_t )$ along the solution. Consider a solution $(\lambda_t ,z_t), t \geq 0$ and the set \(S:=\{(\lambda,z)\mid {\cal L}_f V(\lambda ,z)=0\}=\{(\lambda,z)\mid \lambda =0\}\). Suppose that the solution, $(\lambda_t ,z_t), t \geq 0$, is inside $S$, i.e., $(\lambda_t ,z_t)\in S, t \geq 0$. Then, we should have $\lambda  \equiv 0$, which implies from~(\ref{eq:app:sys:lambda_coordinate_transform}) that $z \equiv 0$. Therefore, $S$ can only contain the trivial solution $(\lambda,z)\equiv (0,0)$. Therefore, from LaSall'es invariance principle in~\Cref{lem:lasalle} and noting that \(V_c\) is radially unbounded, the closed-loop system admits the origin as a globally asymptotically stable equilibrium point. Using the relation $z_t:= x_t- \eta \lambda_t $, we can also easily conclude that the closed-loop system in the original coordinate $(\lambda_t ,x_t)$ admits the origin as a globally asymptotically stable equilibrium point.
\end{proof}

\subsubsection{Proof of~\Cref{lem:st_tdc}}\label{app:lem:st_tdc}
\begin{proof}
One simple option is to set the virtual controller \(\tilde{x}(\lambda_t) :=0 \), which would result to GTD2 as in~\Cref{subsec:btd}. Instead, we take the virtual controller as \( \tilde{x}(\lambda_t) := \beta \lambda_t \), and plug into \(x_t\) in~(\ref{eq:tdc_ode}), which results to
\begin{align}
    \dot{\lambda}_t = - C \lambda_t - \beta A\lambda_t = (- C -\beta A) \lambda_t .\nonumber
\end{align}
The above system is globally asymptotically stable since
\begin{align}
    - C - \beta A \prec 0 \nonumber,
\end{align}
which results from the condition on \(\beta\) in~(\ref{cond:beta}). Using change of coordinates, we introduce the new variable \(z_t\),
\begin{align*}
    z_t := x_t - \tilde{x}(\lambda_t) = x_t - \beta \lambda_t.
\end{align*}
The O.D.E. in~(\ref{eq:tdc_ode}) and~(\ref{eq:tdc_ode_2}) can be rewritten as
\begin{align}
    \dot{\lambda}_t &= (- C - \beta A)\lambda_t - Az_t \label{eq:tdc_ode_z} \\
    \dot{z}_t &= u_t+  (\beta C + \beta^2 A)\lambda_t + \beta Az_t \nonumber.
\end{align}

Now consider the following candidate Lyapunov function:
\begin{align*}
    V_c(\lambda_t,z_t) = \frac{\beta}{2}||\lambda_t||^2 + \frac{1}{2} ||z_t||^2.
\end{align*}
The time derivative of the Lyapunov function along the system's solution becomes,
\begin{align*}
        \dot{V}_c &=  \beta\lambda_t^T ((- C - \beta A)\lambda_t - Az ) + z^{\top}_t (u_t+  (\beta C + \beta^2 A)\lambda_t + \beta Az_t)\\
            &= \beta\lambda_t^T ((- C - \beta A)\lambda_t) + z^{\top}_t(u_t-\beta A^T\lambda_t +(\beta C + \beta^2 A)\lambda_t + \beta Az_t).
\end{align*}

By taking \(u_t = \beta A^{\top}\lambda_t -(\beta C + \beta^2 A)\lambda_t - \beta Az_t \), the corresponding closed-loop system is \( \frac{d}{dt} \begin{bmatrix} 
\lambda_t \\ z_t
\end{bmatrix}
 = \begin{bmatrix}
 -C-\beta A & - A\\
\beta A^{\top} & 0
 \end{bmatrix}
 \begin{bmatrix}
 \lambda_t \\
 z_t
 \end{bmatrix}
 := f(\lambda_t,z_t)
\), and we have \( \dot{V}_c = - \beta ||\lambda_t||^2_{-C+\beta A} \leq 0\) for all \( (\lambda_t,z_t) \neq (0,0) \). We will again use LaSall'es invariance principle in~\Cref{lem:lasalle} in Appendix. Define the Lie derivative \(
\mathcal{L}_f V (\lambda, z) := - \beta ||\lambda||^2_{C+\beta A}
\) so that \(\dot{V}_c (\lambda_t,z_t) = \mathcal{L}_fV(\lambda_t,z_t) \) along the solution. Consider a solution \( (\lambda_t,z_t)\), \(t \geq 0 \) and the set \(
S:= \{(\lambda_t,z_t)|\mathcal{L}_f V(\lambda,z)  =0 \} = \{(\lambda,z)|\lambda=0\}
\). Suppose that the solution \( (\lambda_t,z_t), t\geq 0 \), is inside \(S\), i.e., \( (\lambda_t,z_t)\in S, t\geq 0 \). Then, we should have \(\lambda \equiv 0 \), wchih implies from~(\ref{eq:tdc_ode_z}) that \( z\equiv 0 \). Therefore, \(S\) can only contain the trivial solution \((\lambda,z) \equiv (0,0)\). Therefore, from LaSall'es invariance principle in~\Cref{lem:lasalle} and noting that \(V_c\) is radially unbounded, the closed-loop system admits the origin as a globally asymptotically stable equilibrium point. Using the relation \(z_t = x_t - \beta \lambda_t\), we can also conclude that the closed-loop system in the original coordinate \( (\lambda_t,x_t)\) admits the origin as a globally asymptotically stable equilibrium point.


\end{proof}
Using the relation \(z_t: = x_t - \beta \lambda_t\) , the control input in the original coordinate \( (\lambda_t,x_t )\) can be written as
\begin{align*}
    u_t &=\beta A^{\top}\lambda_t -(\beta C + \beta^2 A)\lambda_t - \beta Az_t\\
      &= \beta A^{\top} \lambda_t - (\beta C + \beta^2 A) \lambda_t -  \beta A (x_t -\beta \lambda_t) \\
      &= (\beta A^{\top}- \beta C) \lambda_t - \beta Ax_t.
\end{align*}
Plugging this input into the original open-loop system in~(\ref{eq:tdc_ode}) and~(\ref{eq:tdc_ode_2}), the closed-loop system in the original coordinate \((\lambda_t,z_t)\) can be written as 
\begin{align}
   \dot{\lambda}_t &= -C  \lambda_t - Ax_t, \label{tdc:ode_lambda}\\
    \dot{x}_t &=  \beta(A^{\top}  -\eta  C )\lambda_t- \beta Ax_t, \label{tdc:ode_x}
\end{align}
whose origin is also globally asymptotically stable according to~\Cref{lem:st_tdc}. Recovering back from \(x_t\) to \(\xi_t\), we have
\begin{align*}
   \dot{\lambda}_t &= -C  \lambda_t - A\xi_t  + b, \nonumber\\
    \dot{\xi}_t &=  \beta(A^{\top}  -\eta  C )\lambda_t- \beta A\xi_t +\beta b,
\end{align*}
whose corresponding stochastic approximation is~(\ref{eq:new_tdc_faster}) and~(\ref{eq:new_tdc_slower}).

\subsubsection{Proof of~\Cref{lem:tdc++ode}}\label{app:lem:tdc++ode}
\begin{proof}
The proof is similar to~\Cref{lem:btd_ode}, hence we breifly explain the procedure. Let the virtual controller \( \tilde{x}(\lambda_t) = \kappa \lambda_t \). We first need to check \(-\eta C - \kappa \eta A - \eta \beta I \) is negative definite. From the condition that \( \beta + \kappa \lambda_{\min} (A)  > \lambda_{\min} (C)\),  \(-\eta C - \kappa \eta A -\eta \beta I \) becomes negative definite.

Using coordinate transform , we define the new variable \(z_t\),
\begin{align}
    z_t = x_t - \tilde{x}(\lambda_t) = x_t - \kappa \lambda_t \nonumber.
\end{align}
Expressing~(\ref{eq:tdc+_ode}) and~(\ref{eq:tdc+u}) in \((\lambda_t,z_t)\), we have
\begin{align*}
    \dot{\lambda}_t &= - \eta (C + \beta I +\kappa  A)\lambda_t - \eta Az_t  \nonumber\\
    \dot{z}_t &= u_t + \eta\kappa  (C + \beta I + \kappa A)\lambda_t + \eta \kappa  Az_t \nonumber.
\end{align*} 
Now, consider the following positive definite function \(V(\lambda_t,z_t) \), and its time derivative:
\begin{align}
    V(\lambda_t,z_t) &= \frac{1}{2\eta}||\lambda_t||^2_2 + \frac{1}{2}||z_t||^2_2 \nonumber,\\
    \dot{V} &= -\lambda^{\top}_t ( C + \beta I +\kappa  A)\lambda_t -  Az_t ) + (z^{\top}_t)(u_t + \eta \kappa (C + \beta I+ \kappa A)\lambda_t+ \kappa \eta Az_t )\nonumber \\
    &= - \lambda^{\top}_t  (C + \beta I +\kappa  A)\lambda_t + (z^{\top}_t)( -A^{\top}\lambda_t + u_t + \kappa \eta (C + \beta I+ \kappa A)\lambda_t + \kappa  \eta Az_t ) \nonumber
\end{align}

Taking \(u_t = A^{\top}\lambda_t -\eta \kappa  (C + \beta I +  \kappa  A)\lambda_t -\kappa  \eta Az_t  \), the corresponding closed-loop system becomes \( 
\frac{d}{dt}\begin{bmatrix}
\lambda_t \\ z_t 
\end{bmatrix} = \begin{bmatrix}
-\eta (C+\beta I + \kappa A) & -\eta 
A\\
A^{\top}  & 0
\end{bmatrix}
\begin{bmatrix}
\lambda_t\\
z_t
\end{bmatrix}
:= f(\lambda_t,z_t)
\), and we have \( \dot{V}_c = - ||\lambda_t||^2_{C+\beta I +\kappa A} \leq 0 \). To use LaSall'es invariance principle in~\Cref{lem:lasalle} given in Appendix, first define the Lie derivative \(\mathcal{L}_f V(\lambda,z):= - ||\lambda||^2_{C+\beta I + \kappa A}\) along the solution. Consider the solution \( (\lambda_t,z_t), t \geq 0 \) and the set \(S:= \{ (\lambda,z) | \mathcal{L}_f V(\lambda,z) = 0 \} = \{ (\lambda,z) | \lambda =0 \}   \). Suppose that the solution \( (\lambda_t,z_t),t\geq 0 \) is inside \(S\), i.e., \( (\lambda_t,z_t)\in S , t \geq 0\). Therefore, \(S\) can only contain the trivial solution \( (\lambda,z) \equiv (0,0) \). Applying Lasall'es invariance principle in~\Cref{lem:lasalle}, we can conclude that the original coordinate \((\lambda_t,x_t)\) admits the origin as globally asymptotically stable equilibrium point. 
\end{proof}

Using the relation \(z_t:= x_t -\kappa \lambda_t\), the control input in the original coordinate \((\lambda_t,x_t)\) can be written as 
\begin{align*}
    u_t &= A^T\lambda_t -\eta \kappa  (C + \beta I + \kappa A)\lambda_t -\kappa  \eta Az_t \\
    &=A^T\lambda_t -\eta\kappa (C + \beta I + \kappa A)\lambda_t - \kappa \eta A(x_t- \kappa \lambda_t)\\
    &= (A^T -\kappa \eta \beta I -\kappa  \eta C)\lambda_t - \kappa \eta Ax_t.
\end{align*}
Plugging this input into the original open-loop system in~(\ref{eq:tdc+_ode}) and~(\ref{eq:tdc+u}), the closed-loop system in the original coordinate \((\lambda_t,x_t) \) can be written as
\begin{align}
    \dot{\lambda}_t &=   -\eta ( C + \beta I)\lambda_t  - \eta A x_t \nonumber\\
 \dot{x}_t &=  (A^{\top}- \kappa \eta(C + \beta I)) \lambda_t - \kappa \eta Ax_t\nonumber.
\end{align}
Recovering back from \(x_t\) to \(\xi_t\) we have
\begin{align}
    \dot{\lambda}_t &=   -\eta ( C + \beta I)\lambda_t  - \eta A \xi_t  + b\nonumber\\
 \dot{\xi}_t &=  (A^{\top}- \kappa \eta(C + \beta I)) \lambda_t - \kappa \eta A\xi_t + \kappa \eta b\nonumber,
\end{align}
whose corresponding stochastic approximation is~(\ref{eq:tdc++2:lambda_update}) and~(\ref{eq:tdc++2:xi_update}).

%% file: algorithms/tdc2.tex
In~\Cref{sec:ss_tdc}, we discussed turning TDC into single-time scale algorithm reflecting the two-time scale property. Other than multiplication of constant reflecting two-time scale property, we can make TDC into single-time scale algorithm as follows:
\begin{align}
    \lambda_{k+1} &= \lambda_k + \alpha_k (-\eta \phi^{\top}_k\lambda_k +\rho_k \delta_k(\xi_k)) \phi_k ) \label{eq:tdc2_lambda}\\
    \xi_{k+1} &= \xi_k+ \alpha_k  (( \phi^{\top}_k \lambda_k \phi_k -\rho_k \gamma \phi^{\top}_k\lambda_k \phi^{\prime}_k) -\eta \phi_k\lambda_k \phi_k +\rho_k \delta_k (\xi_k)\phi_k) \label{eq:tdc2_xi}
\end{align}
 It can be shown that the above stochastic update follows from stabilizing controller design in the following lemma:
\begin{lemma}
Consider the following O.D.E.:
\begin{align}
        \dot{\lambda}_t &=- \eta C\lambda_t - Ax_t \label{eq:tdc2_ode}\\
    \dot{x}_t &= u_t  \nonumber
\end{align}
Suppose we choose the control input \(   u_t:=  (A^{\top}-\eta C) \lambda_t - A x_t  \), and \(\eta\) satisfies condition~(\ref{ineq:meai_eta}). Then, the above O.D.E. has globally asymptotically stable origin, i.e.,
\( (\lambda_t,x_t) \to (0,0)\) as \(t\to \infty\).
\end{lemma}
\begin{proof}
Considering \(x_t\) in~(\ref{eq:tdc2_ode}) as virtual controller \(x_s(\lambda_t) \), one possible option is to take \( x_s(\lambda_t) := 0\). Using backstepping method as in~\Cref{sec:btd}, it results to O.D.E. corresponding to GTD2~\citep{sutton2009fast}. Instead, we choose the virtual controller \( x_s(\lambda_t) := \lambda_t \). Substituting \( x_t \) with \( x_s(t) \) in~(\ref{eq:tdc2_ode}), we can rewrite~(\ref{eq:tdc2_ode}) as follows:
\begin{align*}
    \dot{\lambda}_t = -\eta C \lambda_t- Ax_s(t) = (- \eta C - A) \lambda_t.
\end{align*}
Since \( -\eta C - A \prec 0 \) due to the condition on \(\eta\) in~(\ref{ineq:meai_eta}), the origin of the above system is globally asymptotically stable.
Now, denote the error between the virtual controller \(x_s(t)\) and \(x_t\) as \( z_t := x_t - x_s(\lambda_t) = x_t - \lambda_t \) . Using change of coordinates, we can rewrite the O.D.E. in~(\ref{eq:tdc2_ode}) as follows:
\begin{align}
    \dot{\lambda}_t &= (-\eta C - A)\lambda_t - Az_t \label{eq:tdc2_ode_z} \\
    \dot{z}_t &= u_t +  (\eta C + A)\lambda_t + Az_t \nonumber
\end{align}
Now, let us define the following positive definite function:
\begin{align*}
    V(\lambda_t,z_t)&= \frac{1}{2}||\lambda_t||^2 + \frac{1}{2} ||z_t||^2
\end{align*}
The time derivative of the above function becomes,
\begin{align*}
        \dot{V} &= \lambda^{\top}_t \left((-\eta C - A)\lambda_t - Az_t\right) + z^{\top}_t (u_t +  (\eta C + A)\lambda_t + Az_t)\\
            &= \lambda^{\top}_t (-\eta C - A)\lambda_t + z^{\top}_t(u_t-A^{\top}\lambda_t +(\eta C + A)\lambda_t + Az_t).
\end{align*}
Take \(u_t:=A^{\top}\lambda_t - (\eta C +A)\lambda_t - Az_t\),
the corresponding closed-loop system is 
\( \frac{d}{dt} \begin{bmatrix} 
\lambda_t \\ z_t
\end{bmatrix}
 = \begin{bmatrix}
 -\eta C- A & - A\\
 A^{\top} & 0
 \end{bmatrix}
 \begin{bmatrix}
 \lambda_t \\
 z_t
 \end{bmatrix}
 := f(\lambda_t,z_t)
\), and we have \( \dot{V}_c = - ||\lambda_t||^2_{\eta C + A} \leq 0 \). To use LaSall'es invariance principle in~\Cref{lem:lasalle} given in Appendix, first define the Lie derivative \(\mathcal{L}_f V(\lambda,z):= - ||\lambda||^2_{\eta C + A}\) along the solution.
Consider the solution \( (\lambda_t,z_t),t\geq 0 \) and the set \(S:= \{ (\lambda,z) | \mathcal{L}_f V(\lambda,z) = 0 \} = \{ (\lambda,z) | \lambda =0 \}   \). Suppose that the solution \( (\lambda_t,z_t),t\geq 0 \) is inside \(S\), i.e., \( (\lambda_t,z_t)\in S , t \geq 0\). Therefore, \(S\) can only contain the trivial solution \( (\lambda,z) \equiv (0,0) \). Applying Lasall'es invariance principle in~\Cref{lem:lasalle}, we can conclude that the original coordinate \((\lambda_t,x_t)\) admits the origin as globally asymptotically stable equilibrium point.
\end{proof}

Using the relation \(z_t:= x_t - \lambda_t\), the control input in the original coordinate \((\lambda_t,x_t)\) can be written as 
\begin{align*}
    u_t &= A^{\top}\lambda_t - (\eta C +A)\lambda_t - Az_t \\
    &=A^{\top}\lambda_t - (\eta C +A)\lambda_t - A(x_t-\lambda_t)\\
    &= (A^{\top}-\eta C)\lambda_t - A x_t .
\end{align*}
Plugging this input into the original open-loop system in~(\ref{eq:tdc2_ode}), the closed-loop system in the original coordinate \((\lambda_t,x_t) \) can be written as
\begin{align}
    \dot{\lambda}_t &=   -\eta C \lambda_t -Ax_t \nonumber\\
 \dot{x}_t &=  (A^{\top}-\eta C)\lambda_t -Ax_t \nonumber.
\end{align}
Recovering back from \(x_t\) to \(\xi_t\) we have
\begin{align}
    \dot{\lambda}_t &=   -\eta C \lambda_t -A\xi_t +b \nonumber\\
 \dot{\xi}_t &=  (A^{\top}-\eta C)\lambda_t -A\xi_t +b \nonumber,
\end{align}
whose corresponding stochastic approximation is~(\ref{eq:tdc2_lambda}) and~(\ref{eq:tdc2_xi}).

\begin{remark}
The difference from the update in~\Cref{algo:tdc} can be seen in their corresponding O.D.E. respectively. Multiplying large constant \(\eta\) to \(C\) is enough to make the origin of the O.D.E. stable. 
\end{remark}

By Borkar and Meyn theorem in~\Cref{borkar_meyn_lemma}, we can readily prove the convergence of~\Cref{algo:tdc2}.
\begin{theorem}\label{thm:tdc_slow}
Consider~\Cref{algo:tdc2} given in the Appnedix, which uses stochastic recursive update~(\ref{eq:tdc2_lambda}) and~(\ref{eq:tdc2_xi}). Under the step-size condition~(\ref{cond:robbins_monro_step_size}) and if \(\eta\) satisfies~(\ref{ineq:meai_eta}), then \(\xi_k \to \xi^*\) as \( k \to \infty \) with probability one, where \(\xi^*\) is the TD fixed point in~(\ref{eq:td_fixed_point}).
\end{theorem}

\begin{algorithm}[h]
\caption{Single-time scale TDC2}
  \begin{algorithmic}[1]
    \State Initialize $\xi_0,\lambda_0 \in {\mathbb R}^{n}$.
    \State Set the step-size $(\alpha _k )_{k = 0}^\infty$, and the behavior policy $\mu$.
    \For{iteration $k=0,1,\ldots$}
        \State Sample $s_k\sim d^{\mu}$ and $a_k \sim \mu$
        \State Sample $s_k'\sim P(s_k,a_k,\cdot)$ and $r_k= r(s_k,a_k,s_k')$
        \State Update \(\lambda_k\) and \(\xi_k\) using~(\ref{eq:tdc2_lambda}) and~(\ref{eq:tdc2_xi}) respectively  \EndFor
  \end{algorithmic}\label{algo:tdc2}
\end{algorithm}

Comparison on performance on versions of TDC are given in~\Cref{app:tdc_exp}.

%% file: algorithms/nonlinear.tex
This section provides replacing the so-called regularization term in TDC++ with nonlinear term including ReLU and Leaky ReLU. The basic motiviation is that treating the regularization term as known disturbance, we can cancel the nonlinear term through backstepping. The condition on nonlinear function \( f:\mathbb{R}^n \to \mathbb{R}^n\) are as follows:
\begin{assumption}
\begin{enumerate}
    \item \(f : \mathbb{R}^n \to \mathbb{R}^n\) is Lipschitz continuous.
    \item For some positive constant \(c>0\),
    \begin{align}
  ||f(\lambda_k)||^2_2 \leq  c ||\lambda_k||^2_2 \nonumber.
\end{align}
  \item \(f\) is zero at origin.
  \item \(\lim_{c\to\infty} f(c\lambda_k)/c \to 0\) or \(\lim_{c\to\infty} f(c\lambda_k)/c \to f(\lambda_k) \).
\end{enumerate}
\end{assumption}

\begin{remark}
Such nonlinear terms include ReLU and Leaky ReLU.
\end{remark}
Overall we can prove that the below stochastic recursive update is convergent:
\begin{align}
  \lambda_{k+1} &= \lambda_k + \alpha_k  \eta(- \phi^{\top}_k\lambda_k +   \rho_k \delta_k(\xi_k))\phi_k- \beta f(\lambda_k)  )  \label{eq:tdc++nonlinear:lambda_update}\\
    \xi_{k+1} &= \xi_k+ \alpha_k  (-\rho_k \gamma \phi^{\top}_k\lambda_k \phi^{\prime}_k+(1-\kappa \eta)\phi_k^{\top}\lambda_k\phi_k - \kappa \beta \eta f(\lambda_k) + \kappa \rho_k \eta\delta_k(\xi_k)\phi_k)  \label{eq:tdc++nonlinear:xi_update},
\end{align}
where \(f_k:= f(\lambda_k) \). It has the following corresponding O.D.E. form:
\begin{align}
    \dot{\lambda}_t &=   -\eta C\lambda_t -\eta\beta d_t - \eta A x_t \nonumber\\
 \dot{x}_t &=  (A^{\top}- \kappa \eta C ) \lambda_t -\kappa \eta\beta f(\lambda_t) - \kappa \eta Ax_t\nonumber.
\end{align}

The global asymptotic stability of the above O.D.E. is stated in the following lemma:
\begin{lemma}
Consider the following O.D.E.:
\begin{align}
    \dot{\lambda}_t &= -\eta C \lambda_t -\eta\beta f(\lambda_t) -\eta Ax_t \label{eq:tdcn++ode}\\
    \dot{x}_t &= u_t \label{eq:tdcn++u}.
\end{align}
Suppose we choose the control input \(   u_t:=  (A^T - \kappa\eta \beta f(\lambda_t)- \eta C)\lambda_t - \kappa\eta Ax_t \). Assume \(\eta>0\) and \(\kappa\) and \(\beta\) satisfies the following condition:
\begin{enumerate}
    \item \begin{align}
\begin{cases}\label{cond:nonlinear_kappa}
    0<   \kappa < - \lambda_{\min}(C)\lambda_{\min} \left(\frac{A+A^{\top}}{2}\right) &\text{if} \quad \lambda_{\min}\left( \frac{A+A^{\top}}{2} \right) < 0\\
    0< \kappa & \text{if}\quad \lambda_{\min}\left( \frac{A+A^{\top}}{2} \right) \geq 0
\end{cases}         
\end{align}
\item \begin{align}\label{ineq:tdcn++beta}
    \beta < \frac{1}{c} \lambda_{\max}(C+\kappa A)
\end{align}
\end{enumerate}

Then, the above O.D.E. has globally asymptotically stable origin, i.e.,
\( (\lambda_t,x_t) \to (0,0)\) as \(t\to \infty\).
\end{lemma}

\begin{proof}
The proof is similar to~\Cref{lem:btd_ode}, hence we briefly explain the procedure. Let the virtual controller \( \tilde{x}(\lambda_t) = \kappa \lambda_t \). We first need to check that
\begin{align}
    \dot{\lambda}_t = -\eta C \lambda_t - \eta \beta f(\lambda_t)- \eta \kappa A\lambda_t \nonumber
\end{align}
has globally asymptotically stable origin. Consider the candidate Lyapunov function
\begin{align}
    V (\lambda_t) = \frac{||\lambda_t||^2_2}{2}, \nonumber
\end{align}
which leads to
\begin{align}
    \dot{V}(\lambda_t) = -\eta (C+\kappa A)||\lambda_t||_2^2 - \eta \beta f(\lambda_t)^{\top}\lambda_t \leq -\eta (C+ \kappa A - c \beta I) ||\lambda_t||^2_2 \nonumber.
\end{align}
 \(\dot{V}\) becomes negative definite function due to~(\ref{cond:nonlinear_kappa}) and~(\ref{ineq:tdcn++beta}).

Now, using coordinate transform , we define the error variable \(z_t\),
\begin{align}
    z_t = x_t -  \tilde{x}(\lambda_t) = x_t - \kappa \lambda_t \nonumber.
\end{align}
Expressing~(\ref{eq:tdc+_ode}) and~(\ref{eq:tdc+u}) in \((\lambda_t,z_t)\), we have
\begin{align*}
    \dot{\lambda}_t &= - \eta (C  +\kappa  A)\lambda_t -\eta\beta f(\lambda_t) - \eta Az_t  \nonumber\\
    \dot{z}_t &= u_t + \kappa \eta (C  + \kappa A)\lambda_t + \kappa\eta\beta f(\lambda_t) + \kappa \eta Az_t \nonumber.
\end{align*} 
Now, consider the following positive definite function \(V(\lambda_t,z_t) \), and its time derivative:
\begin{align*}
    &V(\lambda_t,z_t) \nonumber\\
    &= \frac{1}{2\eta}||\lambda_t||^2_2 + \frac{1}{2}||z_t||^2_2 \nonumber\\
    &\dot{V} \nonumber\\
    &= \lambda^{\top}_t (-  (C + \kappa  A)\lambda_t - \beta f(\lambda_t)  - Az ) + (z^{\top}_t)(u_t + \kappa \eta (C  + \kappa A)\lambda_t + \kappa\eta\beta f(\lambda_t) + \kappa \eta Az_t  )\nonumber \\
    &= - \lambda^{\top}_t  ((C + \kappa A)\lambda_t - \beta f(\lambda_t) ) + (z^{\top}_t)( -A^{\top}\lambda_t + u_t + \kappa \eta (C  + \kappa A)\lambda_t + \kappa\eta\beta f(\lambda_t) + \kappa \eta A z_t ) \nonumber
\end{align*}

To achieve \( \dot{V}\leq 0 \), we can choose \(u_t\) as follows:
\begin{align*}
    u_t &= A^{\top}\lambda_t -\kappa \eta (C + \beta f(\lambda_t)+ A)\lambda_t - \kappa\eta Az_t \\
    &=A^{\top}\lambda_t -\kappa\eta (C + \beta f(\lambda_t)+ A)\lambda_t - \kappa\eta A(x_t-\lambda_t)\\
    &= (A^{\top} - \kappa \eta C)\lambda_t - \kappa\eta \beta f(\lambda_t) - \kappa\eta Ax_t
\end{align*}
Using Lasall'es Invariance Principle in~\Cref{lem:lasalle} and similar arguments as before, we can show that the origin is globally asymptotically stable. The proof is complete.
\end{proof}

By Borkar and Meyn theorem in~\Cref{borkar_meyn_lemma}, we can readily prove the convergence of~\Cref{algo:tdc++2}.
\begin{theorem}\label{thm:tdc++2}
Consider~\Cref{algo:tdc++2}. Under the step-size condition~(\ref{cond:robbins_monro_step_size}) and if \(\eta\) satisfies~(\ref{ineq:meai_eta}), then \(\xi_k \to \xi^*\) as \( k \to \infty \) with probability one, where \(\xi^*\) is the TD fixed point in~(\ref{eq:td_fixed_point}).
\end{theorem}

\begin{algorithm}[h]
\caption{TDC++ with nonlinear terms}
  \begin{algorithmic}[1]
    \State Initialize $\xi_0,\lambda_0 \in {\mathbb R}^{n}$.
    \State Set the step-size $(\alpha _k )_{k = 0}^\infty$, and the behavior policy $\mu$.
    \For{iteration $k=0,1,\ldots$}
        \State Sample $s_k\sim d^{\mu`}$ and $a_k \sim \mu$
        \State Sample $s_k'\sim P(s_k,a_k,\cdot)$ and $r_{k+1}= r(s_k,a_k,s_k')$
        \State Update \(\lambda_k\) and \(\xi_k\) using~(\ref{eq:tdc++nonlinear:lambda_update}) and~(\ref{eq:tdc++nonlinear:xi_update}) respectively  \EndFor
  \end{algorithmic}\label{algo:tdc++2}
\end{algorithm}

Here, we present experimental results on TDC++ with nonlinear terms. As shown in below, replacing simple regularization term \( \lambda_k \) with Relu function increases the performance. We set \( \kappa=1,\eta=1,\beta=1 \) and step-size as 0.01. When \(f\) is Relu and LeakyRelu, we call it TDCRelu and TDCLeaky respectively.

\begin{table}[H]
\caption{Best case comparison}
\begin{tabular}{c c c c}\hline
\backslashbox{Env}{Algorithms}
&\makebox[3em]{TDCRelu} &\makebox[3em]{TDCLeaky}&\makebox[3em]{TDC++}\\\hline
Boyan &$\bm{1.392 \pm 0.558}$ & $1.423 \pm 0.55$ & $2.381 \pm 0.256$\\\hline
Dependent &$\bm{0.138 \pm 0.143}$ & $0.139 \pm 0.143$ &$0.201 \pm 0.192$\ \\\hline
Inverted &$\bm{0.358 \pm 0.21}$ &$0.36 \pm 0.211$& $0.493 \pm 0.243$\\\hline
Tabular &$\bm{0.178 \pm 0.174}$& $0.179 \pm 0.175$ & $0.25 \pm 0.245$ \\\hline
Baird & $\bm{0.078 \pm 0.624}$ & $\bm{0.078 \pm 0.624}$ & $0.087 \pm 0.627$ \\\hline
\end{tabular}
\hfill
\centering
\end{table}


%% file: appendix/experiments/main.tex
\subsection{Experiment environments}\label{app:exp_env}
\import{appendix/experiments}{env}

\subsection{Experiment on BTD}\label{app:exp_btd}
\import{appendix/experiments}{exp_btd}

\subsection{Experiment on versions of TDC++}\label{app:exp_tdc++}

\import{appendix/experiments}{exp_tdc++}


%% file: appendix/experiments/env.tex
\subsubsection{Baird counter-example}

Baird's counter-example~\citep{baird1995residual} is a well-known example where TD-learning diverges with over-parameterized linear function approximation. The environment consists of seven states. There are two actions for each state, namely solid and dash action. Solid action leads to the seventh state deterministically, and dash action leads to state other than seventh state with probability 1/6 and no transition occurs with probability 5/6. The behavior policy selects dashed action with probability 1/7, and solid action with 6/7. As in~\cite{baird1995residual}, we set the initial parameters as \([1,\dots,10,1] \). The target policy \(\pi\) only selects solid action at every state.

\begin{align}
    \Phi &:= \begin{bmatrix}
                2 &0& 0& 0& 0& 0& 0& 1\\
            0& 2& 0&0& 0& 0& 0& 1\\
            0& 0& 2& 0& 0& 0& 0& 1\\
        0&0&0&2&0&0&0&1\\
            0& 0& 0& 0& 2& 0& 0& 1\\
            0&0&0&0&0& 2& 0& 1\\
            0 & 0 &0& 0& 0& 0& 1& 2
    \end{bmatrix} \in \mathbb{R}^{7 \times 8},\quad P^{\pi}=
    \begin{bmatrix}
    0& 0& 0& 0& 0& 0& 1\\
    0& 0& 0& 0& 0& 0& 1\\
    0& 0& 0& 0& 0& 0& 1\\
    0& 0& 0& 0& 0& 0& 1\\
    0& 0& 0& 0& 0& 0& 1\\
    0& 0& 0& 0& 0& 0& 1\\
    0& 0& 0& 0& 0& 0& 1
    \end{bmatrix} \in \mathbb{R}^{7\times 7} ,
    \nonumber\\
    D^{\mu} &:= \begin{bmatrix}
    \frac{1}{7}&&\\
    &\ddots&\\
    &&\frac{1}{7}
    \end{bmatrix} \in \mathbb{R}^{7 \times 7} , \quad R^{\pi} = \mathbf{0} \in \mathbb{R}^{7}\nonumber,
\end{align}
where \( \mathbf{0}\) denotes zero column vector with appropriate dimension.

\subsubsection{Boyan Chain}

Boyan Chain~\citep{boyan2002technical} has thirteen states with four features and was designed as on-policy problem. There are two actions and except at state one, where the reward is minus two, each action occurs reward minus three. The behavior policy selects each action under same probability. For states one, five, nine, and 13, the feature vectors are [0,0,0,1],[0,0,1,0],[0,1,0,0], and [1,0,0,0] respectively. The other states are averages over its neighbouring states.

\begin{align}
\Phi&:=\begin{bmatrix}
                1&    0&    0&    0   \\
            3/4& 1/4& 0&    0   \\
            1/2&  1/2&  0&    0   \\
            1/4& 3/4& 0&   0   \\
            0&    1&    0&    0   \\
            0&    3/4& 1/4& 0   \\
            0&    1/2&  1/2&  0   \\
            0&    1/4& 3/4& 0   \\
            0 &    0&    1&   0  \\
            0&   0&    3/4& 1/4\\
            0&    0& 1/2&  1/2 \\
            0&    0&    1/4& 3/4\\
            0&    0&   0&    1   
\end{bmatrix}\in\mathbb{R}^{13\times 4}, \nonumber\\
[P^{\pi}]_{ij} &=
\begin{cases}
\frac{1}{2} & \text{if}\quad i=j+1 \quad\text{or}\quad j=i+1\quad\text{for}\quad 1\leq i \leq 11, 1\leq j \leq 13 \\
1 &\quad \text{if} \quad i=12,\quad j=13 \quad \text{or} \quad i=13 ,\quad j=13
\end{cases} \nonumber,
\\
D^{\mu}&:=\begin{bmatrix}
    \frac{1}{13}&&\\
    &\ddots&\\
    &&\frac{1}{13}
    \end{bmatrix} \in \mathbb{R}^{13 \times 13} 
,\nonumber\\
R^{\pi}&:= \begin{bmatrix}
-3 & -3 & \cdots &-3 &-2 & 0
\end{bmatrix}^{\top} \in \mathbb{R}^{13}
\nonumber
\end{align}

\subsubsection{Random Walk}
Random walk~\citep{sutton2009fast} has five states and two terminal states. Zero rewards occurs except at the last state with a plus one reward. The behavior policy \(\mu\) selects each action with same probability. Following~\cite{sutton2009fast}, we use three different representations, namely tabular, dependent, and inverted features. The diagonal terms are all zero, and off-diagonal terms have a value 1/2, i.e., the feature map of the first state becomes \([0,1/2,1/2,1/2,1/2] \). For the dependent feature case, we have
\([1/\sqrt{3},1/\sqrt{3},1/\sqrt{3}] , [1/\sqrt{2},1/\sqrt{2},0],[1/\sqrt{3},1/\sqrt{3},1/\sqrt{3}] ,[0,1/\sqrt{2},1/\sqrt{2}] \) for each state.

\begin{align}
\Phi_{\text{tabular}}&:=
\begin{bmatrix}
0 &0 &0 &0 &0 &0 &0 \\
0 & 1 &0 &0 &0 &0 &0\\
0 & 0 &1 &0 &0 &0 &0\\
0 & 0 &0 &1 &0 &0 &0\\
0 & 0 &0 &0 &1 &0 &0\\
0 & 0 &0 &0 &0 &1 &0\\
0 & 0 &0 &0 &0 &0 &0\\
\end{bmatrix},\quad \Phi_{\text{inverted}} := \begin{bmatrix}
0 & 0 & 0 & 0 & 0 \\
0 & 1/2 & 1/2 & 1/2 & 1/2\\
1/2 & 0 & 1/2 & 1/2 & 1/2\\
1/2 & 1/2 & 0 & 1/2 & 1/2\\
1/2 & 1/2 & 1/2 & 0 & 1/2\\
1/2 & 1/2 & 1/2 & 1/2 & 0\\
0 & 0 & 0 & 0 & 0
\end{bmatrix} , \nonumber \\
\Phi_{\text{dependent}}&:=
\begin{bmatrix}
0 & 0 & 0\\
 1 & 0 & 0 \\
  1/\sqrt{2}& 1/\sqrt{2} & 0 \\
 1/\sqrt{3} & 1/\sqrt{3} & 1/\sqrt{3} \\
 0& 1/\sqrt{2} & 1/\sqrt{2} \\
 0& 0& 1 \\
 0 & 0 & 0
\end{bmatrix} ,\quad 
P^{\pi} := 
\begin{bmatrix}
1 & 0 & 0 & 0 &0 &0 &0\\
0.6 & 0 & 0.4 & 0 &0 &0 &0\\
0 & 0.6 & 0 & 0.4 &0 &0 &0\\
0 & 0 & 0.6 & 0 & 0.4 &0 &0 \\
0 & 0 & 0 & 0.6 & 0 & 0.4 &0 \\
0 & 0 & 0 & 0 & 0.6 & 0 & 0.4  \\
0 & 0 & 0 & 0 & 0 & 0 & 1  
\end{bmatrix} \nonumber.
\end{align}

%% file: appendix/experiments/exp_btd.tex

\begin{table}[H]
\centering
\caption{Backstepping TD, step-size = $0.1$}
\begin{tabular}{|c|c|c|c|c|c|}\hline
\backslashbox{Env}{$\eta$}
&\makebox[1em]{-0.5}&\makebox[1em]{-0.25}&\makebox[1em]{0}
&\makebox[1em]{0.25}&\makebox[1em]{0.5}\\\hline\hline
Boyan &$0.336 \pm 0.475$ &$0.327 \pm 0.466$ &$0.321 \pm 0.457$ &$0.317 \pm 0.45$ &$ \bm{0.317 \pm 0.443}$\\\hline
Dependent &$0.041 \pm 0.073$ &$0.037 \pm 0.065$ &$0.035 \pm 0.06$ &$\bm{0.034 \pm 0.057}$ &$0.036 \pm 0.057$\\\hline
Inverted &$0.05 \pm 0.105$ &$0.044 \pm 0.092$ &$0.041 \pm 0.083$ &$\bm{0.04 \pm 0.079}$ &$0.041 \pm 0.078$\\\hline
Tabular &$0.075 \pm 0.117$ &$0.068 \pm 0.109$ &$0.064 \pm 0.104$ &$\bm{0.063 \pm 0.102}$ &$0.065 \pm 0.101$\\\hline
Baird&$0.07 \pm 0.596$ &$0.063 \pm 0.593$ &$\bm{0.06 \pm 0.594}$ &$0.06 \pm 0.597$ &$0.063 \pm 0.602$\\\hline
\end{tabular}
\hfill
\end{table}

%% file: appendix/experiments/exp_tdc++.tex
\begin{table}[H]
\centering
\caption{new TDC++, step-size = $0.01$, $\eta=1/2$, $\beta=1$}
\begin{tabular}{|c|c|c|c|c|c|}\hline
\backslashbox{Env}{$\kappa$}
&\makebox[1em]{1/8}&\makebox[1em]{1/4}&\makebox[1em]{1/2}
&\makebox[1em]{1}&\makebox[1em]{2}\\\hline\hline
Boyan &$2.404 \pm 0.242$ &$2.4 \pm 0.241$ &$2.393 \pm 0.243$ &$2.382 \pm 0.255$ &$\bm{2.374 \pm 0.302}$\\\hline
Dependent &$0.166 \pm 0.16$ & $\bm{0.163 \pm 0.156}$ &$0.165 \pm 0.156$ &$0.191 \pm 0.182$ &$0.296 \pm 0.27$\\\hline
Inverted &$\bm{0.366 \pm 0.18}$ &$0.369 \pm 0.183$ &$0.39 \pm 0.194$ &$0.478 \pm 0.233$ &$0.726 \pm 0.342$\\\hline
Tabular &$0.217 \pm 0.216$ &$\bm{0.214 \pm 0.209}$ &$0.215 \pm 0.206$ &$0.238 \pm 0.227$ &$0.342 \pm 0.315$\\\hline
Baird &$0.101 \pm 0.681$ &$0.098 \pm 0.679$ &$\bm{0.094 \pm 0.677}$ &$0.096 \pm 0.681$ &$0.127 \pm 0.711$\\\hline
\end{tabular}
\hfill
\end{table}

\begin{table}[H]
\centering
\caption{new TDC++, step-size = $0.01$, $\eta=1$, $\beta=1$}
\begin{tabular}{|c|c|c|c|c|c|}\hline
\backslashbox{Env}{$\kappa$}
&\makebox[1em]{1/8}&\makebox[1em]{1/4}&\makebox[1em]{1/2}
&\makebox[1em]{1}&\makebox[1em]{2}\\\hline\hline
Boyan &$2.402 \pm 0.241$ &$2.398 \pm 0.241$ &$2.391 \pm 0.243$ &$2.381 \pm 0.256$ &$\bm{2.374 \pm 0.306}$\\\hline
Dependent &$0.163 \pm 0.156$ &$\bm{0.163 \pm 0.155}$ &$0.168 \pm 0.161$ &$0.201 \pm 0.192$ &$0.31 \pm 0.285$\\\hline
Inverted &$\bm{0.367 \pm 0.181}$ &$0.373 \pm 0.186$ &$0.4 \pm 0.199$ &$0.493 \pm 0.243$ &$0.746 \pm 0.356$\\\hline
Tabular &$0.213 \pm 0.204$ &$\bm{0.213 \pm 0.203}$ &$0.218 \pm 0.21$ &$0.25 \pm 0.245$ &$0.36 \pm 0.343$\\\hline
Baird &$0.078 \pm 0.625$ &$0.076 \pm 0.623$ &$\bm{0.076 \pm 0.622}$ &$0.087 \pm 0.627$ &$0.122 \pm 0.655$\\\hline
\end{tabular}
\hfill
\end{table}

\begin{table}[H]
\centering
\caption{new TDC++, step-size = $0.01$, $\eta=2$, $\beta=1$}
\begin{tabular}{|c|c|c|c|c|c|}\hline
\backslashbox{Env}{$\kappa$}
&\makebox[1em]{1/8}&\makebox[1em]{1/4}&\makebox[1em]{1/2}
&\makebox[1em]{1}&\makebox[1em]{2}\\\hline\hline
Boyan &$2.401 \pm 0.24$ &$2.397 \pm 0.24$ &$2.39 \pm 0.243$ &$2.38 \pm 0.256$ &$\bm{2.374 \pm 0.307}$\\\hline
Dependent &$\bm{0.163 \pm 0.156}$ &$\bm{0.163 \pm 0.156}$ &$0.17 \pm 0.164$ &$0.206 \pm 0.198$ &$0.317 \pm 0.293$\\\hline
Inverted &$\bm{0.368 \pm 0.182}$ &$0.376 \pm 0.187$ &$0.405 \pm 0.202$ &$0.502 \pm 0.248$ &$0.757 \pm 0.363$\\\hline
Tabular &$\bm{0.213 \pm 0.203}$ &$0.214 \pm 0.205$ &$0.222 \pm 0.216$ &$0.257 \pm 0.256$ &$0.37 \pm 0.358$\\\hline
Baird &$\bm{0.073 \pm 0.609}$ &$\bm{0.073 \pm 0.609}$ &$0.076 \pm 0.61$ &$0.089 \pm 0.618$ &$0.125 \pm 0.651$\\\hline
\end{tabular}
\hfill
\end{table}

%% file: appendix/experiments.tex
\subsubsection{Comparison between TDC}\label{app:tdc_exp}

In this section we compare single time version of TDC suggested by~\cite{maei2011gradient}, which we denote it as TDC-fast for convenience. We call~\Cref{algo:tdc} as TDC-slow, and~\Cref{algo:tdc2} as TDC2. Under step-size set as 0.01 and 0.1, we swept over \(\eta\in [0.01,0.1,0.5,1,2,4] \) and report best performance. The experiment shows that depending on hyper-parameters and step-size, the performance differs. In general, TDC-slow and TDC2 shows better performance than TDC-fast.

\begin{table}[H]
\centering
\caption{Best case comparison, step-size = $0.01$}\label{table:tdc_best}
\begin{tabular}{|c|c|c|c|}\hline
\backslashbox{Env}{Algorithms}
&\makebox[1em]{TDC-fast}&\makebox[1em]{TDC-slow}&\makebox[1em]{TDC2}\\\hline\hline
Boyan &$0.89 \pm 0.637$ &$0.874 \pm 0.615$ & $\bm{0.533 \pm 0.587}$ \\\hline
Dependent & $0.059 \pm 0.088$ &$0.051 \pm 0.083$ &$\bm{0.043 \pm 0.098}$\\\hline
Inverted &$0.084 \pm 0.115$&$\bm{0.074 \pm 0.106}$& $0.077 \pm 0.123$\\\hline
Tabular & $0.095 \pm 0.124$&$0.078 \pm 0.159$ & $\bm{0.069 \pm 0.159}$ \\\hline
Baird & $\bm{0.057 \pm 0.585}$ & $0.074 \pm 0.614$ &$0.074 \pm 0.614$\\\hline
\end{tabular}
\hfill
\end{table}

\begin{table}[H]
\centering
\caption{Best case comparison, step-size = $0.1$}
\begin{tabular}{|c|c|c|c|}\hline
\backslashbox{Env}{Algorithms}
&\makebox[1em]{TDC-fast}&\makebox[1em]{TDC-slow}&\makebox[1em]{TDC2}\\\hline\hline
Boyan &$0.323 \pm 0.439$ &$0.286 \pm 0.321$ & $\bm{0.268 \pm 0.33}$\\\hline
Dependent & $0.031 \pm 0.046$ &$\bm{0.028 \pm 0.053}$  &$0.031 \pm 0.047$\\\hline
Inverted &$\bm{0.032 \pm 0.056}$&$0.032 \pm 0.058$& $0.032 \pm 0.058$\\\hline
Tabular &$0.052 \pm 0.088$  &$\bm{0.05 \pm 0.092}$  & $0.052 \pm 0.088$  \\\hline
Baird & $0.053 \pm 0.607$ & $\bm{0.05 \pm 0.591}$ &$0.052 \pm 0.609$\\\hline
\end{tabular}
\hfill
\end{table}

The full results are given in Appendix~\Cref{app:exp_tdc++_full}.

\subsubsection{Results on versions of TDC}\label{app:exp_tdc++_full}

Here, we give the full results on the experiments on versions of TDC. We marked '-' in the table if the algorithm diverges.

\begin{table}[H]
\centering
\caption{TDC-fast, step-size = $0.01$}
\begin{tabular}{|c|c|c|c|c|c|}\hline
\backslashbox{Env}{$\eta$}
&\makebox[1em]{0.01}&\makebox[1em]{0.1}&\makebox[1em]{0.5}
&\makebox[1em]{1}&\makebox[1em]{2}\\\hline\hline
Boyan &$0.89 \pm 0.637$ &$1.049 \pm 0.546$ &$1.353 \pm 0.511$ &$1.393 \pm 0.557$ &$1.414 \pm 0.586$\\\hline
Dependent &$0.578 \pm 0.318$ &$0.251 \pm 0.177$ &$0.084 \pm 0.101$ &$0.065 \pm 0.09$ &$0.059 \pm 0.088$\\\hline
Inverted &$0.575 \pm 0.337$ &$0.274 \pm 0.179$ &$0.12 \pm 0.128$ &$0.094 \pm 0.12$ &$0.084 \pm 0.115$\\\hline
Tabular &$0.496 \pm 0.211$ &$0.235 \pm 0.172$ &$0.107 \pm 0.139$ &$0.096 \pm 0.13$ &$0.095 \pm 0.124$\\\hline
Baird &- &- &$0.133 \pm 0.752$ &$0.074 \pm 0.614$ &$0.057 \pm 0.585$\\\hline
\end{tabular}
\hfill
\end{table}

\begin{table}[H]
\centering
\caption{TDC-fast, step-size = $0.1$}
\begin{tabular}{|c|c|c|c|c|c|}\hline
\backslashbox{Env}{$\eta$}
&\makebox[1em]{0.01}&\makebox[1em]{0.1}&\makebox[1em]{0.5}
&\makebox[1em]{1}&\makebox[1em]{2}\\\hline\hline
Boyan &$0.533 \pm 0.352$ &$0.338 \pm 0.356$ &$0.327 \pm 0.414$ &$0.324 \pm 0.43$ &$0.323 \pm 0.439$\\\hline
Dependent&$0.314 \pm 0.274$ &$0.059 \pm 0.103$ &$0.033 \pm 0.052$ &$0.031 \pm 0.047$ &$0.031 \pm 0.046$\\\hline
Inverted&$0.321 \pm 0.25$ &$0.075 \pm 0.112$ &$0.036 \pm 0.064$ &$0.032 \pm 0.058$ &$0.032 \pm 0.056$\\\hline
Tabular &$0.343 \pm 0.239$ &$0.092 \pm 0.126$ &$0.053 \pm 0.09$ &$0.052 \pm 0.088$ &$0.053 \pm 0.087$\\\hline
Baird & - &- &$0.064 \pm 0.623$ &$0.054 \pm 0.608$ &$0.053 \pm 0.607$\\\hline
\end{tabular}
\hfill
\end{table}

\begin{table}[H]
\centering
\caption{TDC-slow, step-size = $0.01$}
\begin{tabular}{|c|c|c|c|c|c|}\hline
\backslashbox{Env}{$\beta$}
&\makebox[1em]{0.01}&\makebox[1em]{0.1}&\makebox[1em]{0.5}
&\makebox[1em]{1}&\makebox[1em]{2}\\\hline\hline
Boyan &$2.804 \pm 0.145$ &$2.574 \pm 0.17$ &$1.898 \pm 0.413$ &$1.393 \pm 0.557$ &$0.874 \pm 0.615$\\\hline
Dependent &$0.474 \pm 0.213$ &$0.222 \pm 0.156$ &$0.094 \pm 0.108$ &$0.065 \pm 0.09$ &$0.051 \pm 0.083$\\\hline
Inverted &$0.612 \pm 0.207$ &$0.299 \pm 0.193$ &$0.132 \pm 0.142$ &$0.094 \pm 0.12$ &$0.074 \pm 0.106$\\\hline
Tabular &$0.771 \pm 0.343$ &$0.272 \pm 0.243$ &$0.128 \pm 0.153$ &$0.096 \pm 0.13$ &$0.078 \pm 0.116$\\\hline
Baird &$1.844 \pm 2.12$ &$0.218 \pm 1.011$ &$0.084 \pm 0.665$ &$0.074 \pm 0.614$ &$0.084 \pm 0.638$\\\hline
\end{tabular}
\hfill
\end{table}

\begin{table}[H]
\centering
\caption{TDC-slow, step-size = $0.1$}
\begin{tabular}{|c|c|c|c|c|c|}\hline
\backslashbox{Env}{$\beta$}
&\makebox[1em]{0.01}&\makebox[1em]{0.1}&\makebox[1em]{0.5}
&\makebox[1em]{1}&\makebox[1em]{2}\\\hline\hline
Boyan &$2.588 \pm 0.168$ &$1.433 \pm 0.613$ &$0.452 \pm 0.571$ &$0.324 \pm 0.43$ &$0.286 \pm 0.321$\\\hline
Dependent &$0.223 \pm 0.154$ &$0.058 \pm 0.087$ &$0.028 \pm 0.053$ &$0.031 \pm 0.047$ &$0.042 \pm 0.047$\\\hline
Inverted &$0.295 \pm 0.183$ &$0.08 \pm 0.109$ &$0.032 \pm 0.068$ &$0.032 \pm 0.058$ &$0.041 \pm 0.055$\\\hline
Tabular &$0.273 \pm 0.236$ &$0.097 \pm 0.12$ &$0.05 \pm 0.092$ &$0.052 \pm 0.088$ &$0.064 \pm 0.085$\\\hline
Baird &$0.215 \pm 0.951$ &$0.055 \pm 0.584$ &$0.05 \pm 0.591$ &$0.054 \pm 0.608$ &-\\\hline
\end{tabular}
\hfill
\end{table}

\begin{table}[H]
\centering
\caption{TDC2, step-size = $0.01$}
\begin{tabular}{|c|c|c|c|c|c|}\hline
\backslashbox{Env}{$\eta$}
&\makebox[1em]{0.01}&\makebox[1em]{0.1}&\makebox[1em]{0.5}
&\makebox[1em]{1}&\makebox[1em]{2}\\\hline\hline
Boyan &$0.73 \pm 0.574$ &$0.533 \pm 0.587$ &$0.883 \pm 0.652$ &$1.393 \pm 0.557$ &$1.894 \pm 0.41$\\\hline
DependentRep &$0.051 \pm 0.106$ &$0.043 \pm 0.098$ &$0.045 \pm 0.086$ &$0.065 \pm 0.09$ &$0.104 \pm 0.11$\\\hline
InvertedRep &$0.109 \pm 0.14$ &$0.093 \pm 0.134$ &$0.077 \pm 0.123$ &$0.094 \pm 0.12$ &$0.137 \pm 0.133$\\\hline
TabularRep &$0.083 \pm 0.166$ &$0.069 \pm 0.159$ &$0.076 \pm 0.134$ &$0.096 \pm 0.13$ &$0.133 \pm 0.149$\\\hline
BairdRep &-&- &$0.125 \pm 0.72$ &$0.074 \pm 0.614$ &$0.076 \pm 0.62$\\\hline
\end{tabular}
\hfill
\end{table}

\begin{table}[H]
\centering
\caption{TDC2, step-size = $0.1$}
\begin{tabular}{|c|c|c|c|c|c|}\hline
\backslashbox{Env}{$\eta$}
&\makebox[1em]{0.01}&\makebox[1em]{0.1}&\makebox[1em]{0.5}
&\makebox[1em]{1}&\makebox[1em]{2}\\\hline\hline
Boyan &$0.324 \pm 0.313$ &$0.275 \pm 0.274$ &$0.268 \pm 0.33$ &$0.324 \pm 0.43$ &$0.456 \pm 0.569$\\\hline
DependentRep &$0.049 \pm 0.052$ &$0.043 \pm 0.049$ &$0.034 \pm 0.046$ &$0.031 \pm 0.047$ &$0.032 \pm 0.055$\\\hline
InvertedRep &$0.053 \pm 0.067$ &$0.046 \pm 0.063$ &$0.035 \pm 0.058$ &$0.032 \pm 0.058$ &$0.036 \pm 0.067$\\\hline
Tabular &$0.068 \pm 0.095$ &$0.062 \pm 0.091$ &$0.053 \pm 0.087$ &$0.052 \pm 0.088$ &$0.055 \pm 0.093$\\\hline
Baird & - &-&$4.492 \pm 57.657$ &$0.054 \pm 0.608$ &$0.052 \pm 0.609$\\\hline
\end{tabular}
\hfill
\end{table}

\subsubsection{Overall comparison between the algorithms}

\begin{table}[H]
\centering
\caption{Overall comparison, step-size = $0.01$}
\begin{tabular}{|c|c|c|c|c|c|}\hline
\backslashbox{Algorithms}{Env}
&\makebox[1em]{Boyan}&\makebox[1em]{Dependent}&\makebox[1em]{Inverted}
&\makebox[1em]{Tabular}&\makebox[1em]{Baird}\\\hline\hline
TD & $ 0.536 \pm 0.699$& $ \mathbf{0.016 \pm 0.029}$&$ \mathbf{0.03 \pm 0.041}$ & $0.028 \pm 0.036$ & - \\\hline
GTD2 &$ 1.452 \pm 0.647 $ & $0.086 \pm 0.142$ &$0.151 \pm 0.193$ &$0.133 \pm 0.208$ &$0.085 \pm 0.625 $\\\hline
BTD ($\eta=0.5$)&$1.408 \pm 0.635$ &$0.076 \pm 0.122$ &$0.136 \pm 0.172$ &$0.122 \pm 0.188$ &$0.092 \pm 0.637$\\\hline
TDC-fast &$0.89 \pm 0.637$ &$0.059 \pm 0.088$ &$0.084 \pm 0.115$ &$0.095 \pm 0.124$ &$ \mathbf{0.057 \pm 0.585}$\\\hline
TDC-slow &$ 0.874 \pm 0.615$ &$0.051 \pm 0.083$ &$0.074 \pm 0.106$ &$0.078 \pm 0.159$ &$0.074 \pm 0.614$\\\hline
TDC2 & $0.533\pm 0.587$& $0.043\pm 0.098$ & $0.077 \pm 0.123 $ &$ 0.069 \pm 0.159$ &$0.074 \pm 0.614$\\\hline
ETD & $\mathbf{0.469\pm 0.599}$& $0.019\pm 0.023$ & $0.032 \pm 0.035 $ &$ \mathbf{0.021 \pm 0.028}$ &-\\\hline
\end{tabular}
\hfill
\end{table}

Even though TD and ETD ( Emphatic Temporal-Difference learning)~(\cite{mahmood2015emphatic}) shows good performance in several domains, it shows unstable behavior in Baird's counter example. TDC-fast shows better performance than other algorithms in Baird's counter example, but in other domains it shows worse performance than TDC-slow or TDC2 as can be seen in~\Cref{table:tdc_best}. Moreover, when \( \eta = 0.5 \), BTD shows better performance than GTD2 except at Baird's counter example.

\subsection{O.D.E. results}

In this section, we see how each O.D.E. dyanmics of TD-learning algorithm dynamics behave in Baird counter example.

\begin{figure}[H]
     \begin{subfigure}[t]{0.49\textwidth}
         \centering
         \includegraphics[width=1\textwidth]{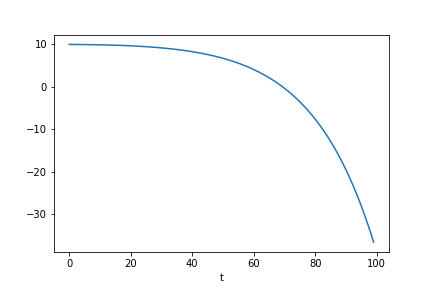}
            \caption{TD}
     \end{subfigure}
          \hfill
              \begin{subfigure}[t]{0.49\textwidth}
         \centering
         \includegraphics[width=1\textwidth]{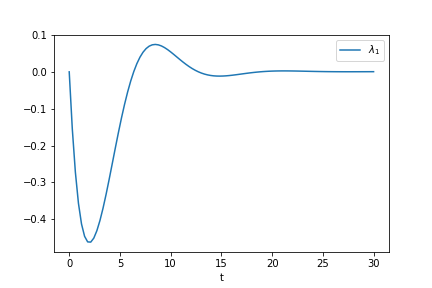}
         \caption{GTD2}
     \end{subfigure}
          \begin{subfigure}[t]{0.49\textwidth}
         \centering
         \includegraphics[width=1\textwidth]{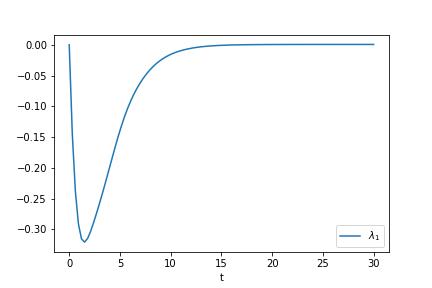}
          \caption{TDC-fast, \(\eta=1\)}
     \end{subfigure}
               \begin{subfigure}[t]{0.49\textwidth}
         \centering
         \includegraphics[width=1\textwidth]{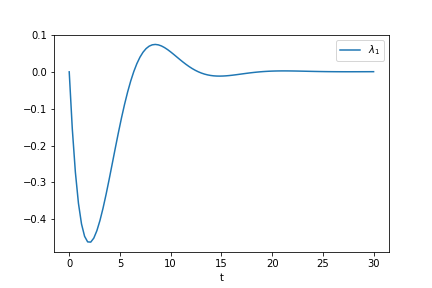}
          \caption{BTD, \(\eta=0.5\)}
     \end{subfigure}
     \caption{O.D.E. dynamics of first element of \( \lambda_t \) in Baird counter example}
\end{figure}

As can be seen in (a), the O.D.E. behavior of TD shows divergence, whereas other algorithms including GTD2, TDC-fast, and BTD show stable behavior. GTD2 and BTD, compared to TDC, show some oscillatory behavior, which may cause slow convergence compared to TDC in Baird counter example.

%% file: appendix/background_additional.tex
~\cite{sutton2009fast} introduced GTD2 and TDC to find the global minimizer of MSPBE:
\begin{align*}
    \frac{1}{2}
    \left\lVert 
    \Phi^{\top}D^{\mu}\Phi
    \theta - \gamma \Phi^{\top} D^{\mu}P^{\pi}\Phi\theta - \Phi^{\top}DR
    \right\rVert^2_{C^{-1}}.
\end{align*}

Taking gradient fo the above equation with respect to \(\theta\), we get
\begin{align*}
(\Phi^{\top}D^{\mu}\Phi
   - \gamma \Phi^{\top} D^{\mu}P^{\pi}\Phi)^{\top}
    C^{-1}\left(
 \Phi^{\top}D^{\mu}\Phi
    \theta - \gamma \Phi^{\top} D^{\mu}P^{\pi}\Phi\theta - \Phi^{\top}DR
    \right).
\end{align*}
The above gradient is equal to \( \mathbb{E}[(\gamma \phi^{\prime}_k-\phi_k)\phi_k^{\top}]\mathbb{E}[\phi_k\phi_k^{\top}]^{-1}\mathbb{E}[(r_k+\gamma \phi^{\prime\top}_k \xi_k - \phi_k^{\top}\xi_k)\phi_k]\), and due to the inverse operation and double sampling issue (dependency of \(\phi^{\prime}_k\) in the first term and last term), stochastic samples would lead to significant biases. Hence, GTD2 and TDC try to approximate the stochastic gradient of MSPBE, i.e., the term \( (\Phi^{\top}D^{\mu}\Phi- \gamma \Phi^{\top} D^{\mu}P^{\pi}\Phi)^{\top}C^{-1}\).